\documentclass[11pt]{article}

\usepackage[preprint]{acl}

\usepackage{times}
\usepackage{latexsym}
\usepackage{url}
\usepackage{algorithm}
\usepackage{algorithmic}
\usepackage{amssymb}
\usepackage{amsmath}
\usepackage{array}
\usepackage{bbm}
\usepackage{bm}
\usepackage{booktabs}
\usepackage{float}
\usepackage{multirow}
\usepackage{bbding}

\usepackage{color}
\usepackage{xspace,mfirstuc,tabulary}

\usepackage[T1]{fontenc}
\usepackage[utf8]{inputenc}

\usepackage{microtype}

\usepackage{inconsolata}

\usepackage{graphicx}

\usepackage{subcaption}
\usepackage{tabularx}
\usepackage{makecell}

\title{Label Semantic Expansion via Label Guided Neural Topic Modeling}

\author{
  Haojia Zheng\textsuperscript{*\P},
  Yuyin Lu\textsuperscript{*\P},
  Juntian Huang\textsuperscript{*},
  Fan Ou\textsuperscript{*},
\\
  Yanghui Rao\textsuperscript{*\ensuremath{\parallel}},
  Haoran Xie\textsuperscript{\ensuremath{\dagger}},
  Fu Lee Wang\textsuperscript{\ensuremath{\ddagger}}
\\[1mm]
  \textsuperscript{*}School of Computer Science and Engineering,
  Sun Yat-sen University, Guangzhou, China
\\
  \textsuperscript{\ensuremath{\dagger}}Division of Artificial Intelligence,
  Lingnan University, Hong Kong SAR, China
\\
  \textsuperscript{\ensuremath{\ddagger}}School of Science and Technology,
  Hong Kong Metropolitan University, Hong Kong SAR, China
\\[1mm]
  \textsuperscript{\P}Equal contribution.
  \qquad
  \textsuperscript{\ensuremath{\parallel}}Corresponding author.
}

\usepackage[most]{tcolorbox}
\usepackage{xcolor}
\tcbset{
  promptbox/.style={
    enhanced,
    colback=white,
    colframe=black,
    boxrule=0.9pt,
    arc=1mm,
    left=1.2mm,
    right=1.2mm,
    top=1mm,
    bottom=1mm,
    fonttitle=\bfseries,
    coltitle=white,
    title filled=false
  }
}

\begin{document}
\maketitle

\begin{abstract}
Topic models are widely used for content analysis, where users often analyze corpora around predefined labels rather than unordered latent topics. 
Existing label-aware topic models mainly follow a \emph{labels-for-topics} perspective, using labels to guide topic learning, while the learned topics are not directly usable for label-centered analysis. 
We explore the reverse \emph{topics-for-labels} perspective and instantiate it as \emph{Label Semantic Expansion} (LSE), which enriches sparse label representations with corpus-grounded descriptive topic words. 
To exploit topics in LSE effectively, we propose a \emph{Label-Guided Neural Topic Model} (LGNTM), which learns dedicated label-aligned topics, grounds them in lexical and document semantic spaces, and preserves consistency between topic structures and label structures. 
Experiments on label-topic alignment, label expansion, topic quality, and downstream classification demonstrate strong overall performance across complementary evaluation dimensions\footnote{Our implementation and all datasets used in the experiments are publicly available at \url{https://anonymous.4open.science/r/LGNTM-2E0C/}.}.
\end{abstract}

% \input{docs_new/ch1-0409}
% \input{docs_new/ch1-0506}
% \input{docs_new/ch1-0522}

% \input{docs/ch2-Related-Work}
% \input{docs_new/ch2-0523}
% \input{docs_new/ch1-0523}
% \input{docs_new/ch1&2-0524}
% \input{docs_0525/ch1&2}
% \section{Introduction}

% % 主题模型是内容分析工具；但内容分析通常不是只需要一组无标签 latent topics，而是需要围绕人类预定义分析类别理解语料。
% % 标签与主题对齐的价值，激发了很多 supervised TM
% % 然而他们是 labels for topics, 最终得到的主题仍然需要后处理，增加负担
% % 他们忽略了 topics for labels 的反向角度
% % topics for labels 其形式化为 LSE，该视角 不仅可直接围绕人类预定义类别主题分析语料（因为每个标签均能直接得到一个基于语料库分析结果的主题），且解决了 label 标签名短信息量不足的问题
% % 举例 table 1 当前模型做不到，简单讲
% % 我们 从 topics for labels 的视角 提出了 LGNTM ...

\section{Introduction}

Topic models have been widely used for content analysis because they can uncover coherent semantic structures from text corpora through topic-word and document-topic distributions~\cite{organizing-text, Event-Analysis, chen2025structural}. 
However, practical exploratory analysis often starts from specific analytical objectives rather than from inspecting all latent topics in a corpus~\cite{TopicSifter}. 
In many real-world applications, such objectives are reflected by human-predefined labels, such as policy areas~\cite{hoyle-are-NTM-broken}, news topics~\cite{20news}, and scientific fields~\cite{wos_key}. 
These labels often capture salient semantic categories of interest to users and can serve as human-defined anchors for content analysis~\cite{label-informative}. 
Accordingly, prior work has emphasized the importance of label-topic alignment for both interpretability and practical content analysis~\cite{hoyle-are-NTM-broken}. 
These observations suggest that the usefulness of learned topics depends not only on their internal semantic quality, but also on whether they can be organized around the predefined categories that users intend to analyze.

In response to this need, many supervised topic models have been developed. 
They follow a \emph{labels-for-topics} perspective, where labels serve as supervision signals for guiding label-topic association and learning better topics, by supporting credit attribution~\cite{LabeledLDA}, improving predictive modeling~\cite{sLDA, scholar}, or enhancing label-topic interpretability~\cite{LANTM}.
However, most of these models still output unlabeled topic-word lists that are not directly organized around human-predefined categories. 
As a result, users need to infer topic meanings and manually match anonymous topics to predefined labels, increasing the burden of post-hoc interpretation~\cite{bhatia-etal-2016-automatic}.

The above limitation suggests that label-topic alignment should not be used to learn better topics only. 
It also raises a reverse question: if labels can guide topic learning, can the learned topics in turn serve as semantic representations of the labels themselves? 
We refer to this reverse perspective as \emph{topics-for-labels}. 
Under this perspective, learned topics are expected to support category-driven analysis by organizing corpus-level semantic evidence around human-predefined labels, rather than leaving users to interpret anonymous topics post hoc. 
This requires each predefined label to be associated with a dedicated topic-word distribution.

We instantiate the topics-for-labels perspective as \emph{Label Semantic Expansion} (LSE). 
Given a labeled corpus, LSE aims to enrich each predefined label with corpus-grounded descriptive words extracted from its associated topic-word distribution. 
This formulation addresses the limited informativeness of short category labels: representing a concept with a single term often provides insufficient information about its lexical scope and semantic boundary, and may introduce ambiguity~\cite{label-related3}. 
By expanding short labels into corpus-grounded lexical descriptions, LSE makes predefined categories more explicit and interpretable, further supporting category-driven content analysis. 
Such enriched label semantics can also benefit downstream content analysis tasks, such as label name-based text  classification~\cite{LOTClass} and prompt augmentation~\cite{LAAV, KPT}.

\begin{table}[t]
\centering
\small
\setlength{\tabcolsep}{3pt}
% \caption{A case study of expanded words for the label ``biochemistry'' produced by LLDA, LANTM and LGNTM.}
\caption{Expanded words for the label ``biochemistry'' generated by Labeled LDA (LLDA)~\cite{LabeledLDA}, LANTM~\cite{LANTM}, and our LGNTM.}
\begin{tabular}{@{}lp{0.70\linewidth}@{}}
\toprule
Model & Expansion for ``biochemistry'' \\
\midrule
LLDA & cell, study, use, human, result \\
LANTM & protein, cell, accelerate, advanced, system \\
LGNTM & gene, metabolic, cell, mutation, dna \\
\bottomrule
\end{tabular}
\label{tab:case-study}
\end{table}

% 这段话之前，需要加一下过渡句子，说明LSE这个任务的挑战，再用Table 1作为佐证。
However, LSE is challenging because label-topic association does not necessarily yield label-specific and informative expansion words.
As shown in Table~\ref{tab:case-study}, both Labeled LDA and LANTM produce expansions for ``biochemistry'' that still contain generic or weakly related corpus words.
This suggests that label-topic association alone, whether imposed by hard restriction or encouraged by soft guidance, is insufficient to produce useful label expansions.

To address this gap, we propose a \emph{Label-Guided Neural Topic Model} (LGNTM) to realize the \emph{topics-for-labels} perspective for LSE.
Our model learns a one-to-one correspondence between predefined labels and latent topics, so that each label can be represented by the topic-word distribution of its aligned topic. 
To make these label-aligned topics dedicated, discriminative, and faithful to the corpus, LGNTM combines label-guided topic specialization, dual semantic grounding in lexical and document embedding spaces, and hierarchy-aware consistency regularization.

Our contributions are summarized as follows:
\begin{itemize}
    \item We introduce the \emph{topics-for-labels} perspective and formulate \emph{Label Semantic Expansion} as the task of enriching predefined labels with corpus-grounded descriptive words.
    \item We propose LGNTM, a label-guided topic model that learns dedicated label-aligned topics, grounds them in corpus semantics, and preserves hierarchical label consistency.
    \item We evaluate LGNTM in terms of label-topic alignment, label expansion, topic quality, and downstream classification, demonstrating the utility of learned label-aligned topics.
\end{itemize}

\section{Related Work}

\begin{figure}[t]
    \centering
    \begin{subfigure}{0.31\columnwidth}
        \centering
        \includegraphics[width=\linewidth]{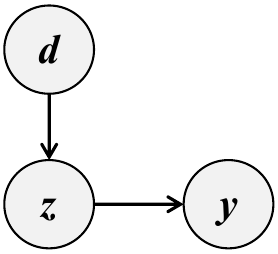}
        % \caption{SCHOLAR}
        \caption{ }
        \label{fig:mode_prediction}
    \end{subfigure}
    \hfill
    \begin{subfigure}{0.31\columnwidth}
        \centering
        \includegraphics[width=\linewidth]{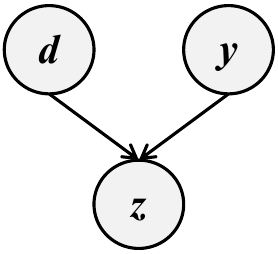}
        % % \caption{LATNM}
        \caption{ }
        \label{fig:mode_label_guided}
    \end{subfigure}
    \hfill
    \begin{subfigure}{0.31\columnwidth}
        \centering
        \includegraphics[width=\linewidth]{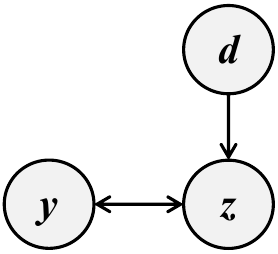}
        % \caption{LGNTM}
        \caption{ }
        \label{fig:mode_topics_for_labels}
    \end{subfigure}
    \caption{Three paradigms of using labels in supervised topic models, where $d$, $z$, and $y$ denote document, latent topics, and predefined labels, respectively.}
    \label{fig:three_modes}
\end{figure}

Supervised topic models incorporate labels into topic learning in different ways. 
As illustrated in Figure~\ref{fig:three_modes}\subref{fig:mode_prediction}, prediction-oriented methods such as sLDA~\cite{sLDA}, SNTM~\cite{SNTM}, and SCHOLAR~\cite{scholar} use document-topic distributions to predict labels, inducing implicit topic-label associations. 
As illustrated in Figure~\ref{fig:three_modes}\subref{fig:mode_label_guided}, label-guided methods introduce labels more directly into topic inference: Labeled LDA~\cite{LabeledLDA} restricts each document to topics associated with its observed labels, while neural supervised topic models use label-conditioned priors, soft indicators, or alignment objectives to encourage label-topic consistency~\cite{LI-NTM, FANToM, LANTM}. 
For example, LANTM~\cite{LANTM} aligns document-topic distributions with label information through label priors and soft label-topic indicators.

Although the above methods improve label-topic association, they differ in how far they satisfy the \emph{topics-for-labels} requirement. 
Prediction-oriented models (e.g., SCHOLAR) still output unlabeled topic-word lists and require post-hoc label matching.
LANTM identifies label-relevant topics with soft indicators, but multiple labels may share dominant topics, and dedicated label-specific topic-word distributions are not guaranteed.
Labeled LDA is structurally closest to topics-for-labels because each label is tied to a topic, but its credit-attribution and label-restricted reconstruction objectives do not guarantee label-specific topic words. 
Overall, existing supervised topic models remain within a \emph{labels-for-topics} paradigm, limiting their ability to directly support content analysis around predefined categories.

% \input{docs/ch3-Method}
% docs_new/ch3&4-0508}
% \input{docs_new/ch3&4-0523}
% \input{docs_new/ch3&4-0524}
% \input{docs_0525/ch3&4}

\section{Background and Problem Setup}
\label{sec:Problem-Formulation}

\paragraph{Background}
Given a document $d$, a topic model infers a document-topic distribution $p(z\,|\,d)$, which describes the association between the document and latent topics. Each topic $z$ is further associated with a topic-word distribution $p(w\mid z)$, which provides a lexical description of the topic.

% $y_d^{(l)}$：表示文档 $d$ 在第 $l$ 层的观测标签。
% $y_k^{(l)}$：表示第 $l$ 层标签集合中的第 $k$ 个标签类别。

% We consider a labeled corpus where each document $d$ has labels $\{y_d^{(1)},\ldots,y_d^{(H)}\}$ across $H$ levels, with $y_d^{(l)}\in\mathcal{Y}^{(l)}=\{1,\ldots,K_l\}$. Flat label structures are a special case with $H=1$.

% We consider a labeled corpus where each document $d$ has labels $\{y_d^{(l)}\}_{l=1}^{H}$, with $y_d^{(l)}\in\mathcal{Y}^{(l)}$ and $|\mathcal{Y}^{(l)}|=K_l$. Flat label structures correspond to $H=1$.

We consider a labeled corpus where each document $d$ has labels $\{y_d^{(l)}\}_{l=1}^{H}$, with $y_d^{(l)}\in\mathcal{Y}^{(l)}$, $|\mathcal{Y}^{(l)}|=K_l$; flat labels have $H=1$.

\paragraph{Topics-for-Labels}

% Following the topics-for-labels perspective, we ask whether label-aligned topics can be used to enrich predefined labels with corpus-grounded topic-words. We refer to this task as \emph{Label Semantic Expansion} (LSE). 
% Given a label $y_k^{(l)}$ at level $l$, LSE aims to produce an expansion word set $\mathcal{E}_{k}^{(l)}$ from a label-conditioned word distribution $p(w\mid y_k^{(l)})$.

Following the topics-for-labels perspective, we ask whether label-aligned topics can enrich predefined labels with corpus-grounded topic words. We refer to this task as \emph{Label Semantic Expansion} (LSE). For label $y_k^{(l)}\in\mathcal{Y}^{(l)}$, LSE produces an expansion word set $\mathcal{E}_{k}^{(l)}$ from $p(w\mid y_k^{(l)})$.

To instantiate this idea, our LGNTM introduces a label-topic correspondence function, as illustrated in Figure~\ref{fig:three_modes}\subref{fig:mode_topics_for_labels}:
\begin{equation}
\Pi_l:\mathcal{Y}^{(l)} \leftrightarrow \mathcal{Z}^{(l)}, \qquad \Pi_l(y_k^{(l)})=z_k^{(l)},
\label{eq:double-shot}
\end{equation}
where $\mathcal{Z}^{(l)}$ is the topic set at level $l$. 
To realize this correspondence, LGNTM sets the number of topics at each level equal to the number of labels, i.e., $|\mathcal{Z}^{(l)}|=K_l$. The label-conditioned word distribution is then parameterized by the topic-word distribution of its corresponding topic:
\begin{equation}
p(w\mid y_k^{(l)}) := p(w\mid \Pi_l(y_k^{(l)})) = p(w\mid z_k^{(l)}).
\label{eq:lse_topic_parameterization}
\end{equation}

\section{Label-Guided Neural Topic Model}
\label{sec:Methodology}

\begin{figure*}[t]
    \centering
    \includegraphics[width=\textwidth]{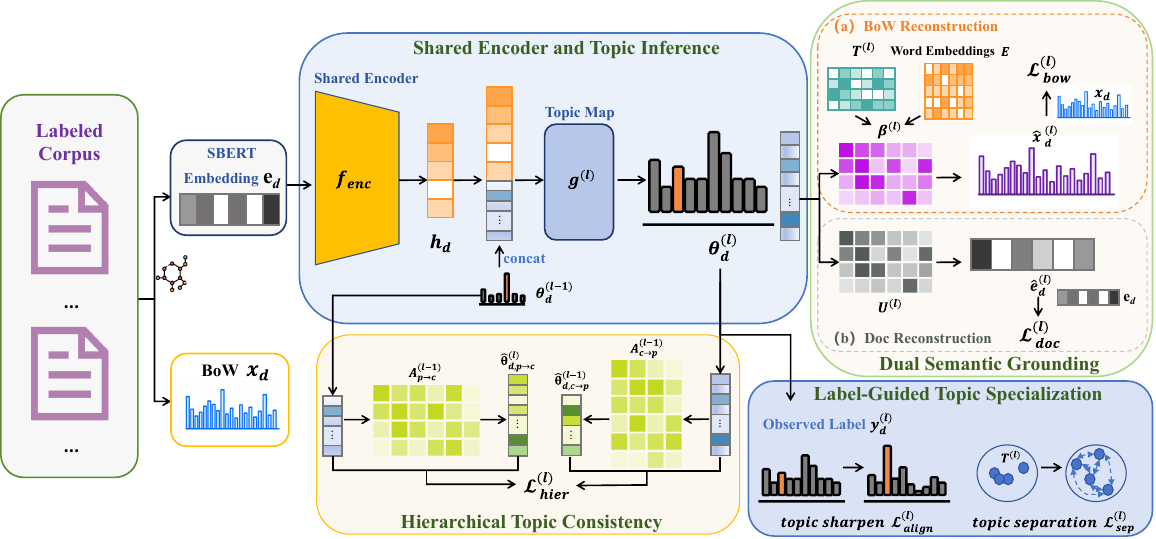}
    \caption{
    Overall architecture of LGNTM. 
    }
    \label{fig:framework}
\end{figure*}

\subsection{Model Overview}

% Based on the topics-for-labels setup in Section~\ref{sec:Problem-Formulation}, LGNTM learns label-specialized topic representations for LSE. It consists of three components: label-guided topic specialization, which assigns each label a dedicated and distinguishable topic; dual semantic grounding, which ties label-specialized topics to corpus-level lexical evidence and contextual document semantics; and hierarchical topic consistency, which preserves compatibility across label levels. Figure~\ref{fig:framework} illustrates the overall architecture with a shared document encoder and a hierarchical topic decoder.

% Each document $d$ is represented by a bag-of-words vector 
% $x_d\in\mathbb{R}^{V}$ and a contextual document embedding 
% $e_d\in\mathbb{R}^{D_e}$. The vocabulary is associated with a word 
% embedding matrix $W\in\mathbb{R}^{V\times D_w}$.

Based on the topics-for-labels setup in Section~\ref{sec:Problem-Formulation}, LGNTM learns label-specialized topic representations for LSE. It consists of three components: label-guided topic specialization, which assigns each label a dedicated and distinguishable topic; dual semantic grounding, which ties label-specialized topics to corpus-level lexical evidence and contextual document semantics; and hierarchical topic consistency, which preserves compatibility across label levels. Figure~\ref{fig:framework} illustrates the overall architecture with a shared document encoder and a hierarchical topic decoder. Each document $d$ is represented by a bag-of-words vector $x_d\in\mathbb{R}^{V}$ and a contextual document embedding $e_d\in\mathbb{R}^{D_e}$, and the vocabulary is associated with a word embedding matrix $W\in\mathbb{R}^{V\times D_w}$.

Given the document embedding $e_d$, the shared encoder produces a hidden representation:
\begin{equation}
h_d = f_{\mathrm{enc}}(e_d).
\end{equation}

% To pass coarse-level topic information to finer levels, the decoder input at label level $l$ is defined as
For coarse-to-fine topic propagation, the decoder input at level $l$ is
\begin{equation}
u_d^{(l)}
=
\begin{cases}
h_d, & l=1,\\
\left[h_d;\theta_d^{(l-1)}\right], & l>1,
\end{cases}
\label{eq:decoder_input}
\end{equation}
where $[\cdot;\cdot]$ denotes concatenation. The document-topic distribution at level $l$ is then computed as
\begin{equation}
\theta_d^{(l)}
=
\mathrm{softmax}
\left(
g^{(l)}(u_d^{(l)})
\right),
\label{eq:theta}
\end{equation}
% where $g^{(l)}(\cdot)$ is a level-specific affine map and $\theta_d^{(l)} \in \Delta^{K_l}$.
where $g^{(l)}(\cdot)$ is an affine map and $\theta_d^{(l)} \in \Delta^{K_l}$.
% where $g_l$ maps $u_d^{(l)}$ to topic logits and $\theta_d^{(l)} \in \Delta^{K_l}$.

Each level maintains a topic embedding matrix $T^{(l)} \in \mathbb{R}^{K_l \times D_w}$, where each row is a topic embedding in the same space as the word embeddings $W$. Following ETM~\cite{ETM}, the topic-word distribution is computed by matching topic and word embeddings:
\begin{equation}
\beta^{(l)}
=
\mathrm{softmax}
\left(
T^{(l)} W^\top
\right),
\label{eq:beta}
\end{equation}
where the softmax function is applied over the vocabulary dimension. 
By introducing the label-topic correspondence in Eq.~\eqref{eq:double-shot}, each row $\beta_k^{(l)}\in\Delta^V$ serves as the label-conditioned word distribution for label $y_k^{(l)}$.

\subsection{Label-Guided Topic Specialization}

Following the label-topic correspondence, LGNTM constructs a label-indexed topic space at each label level. This is achieved through distribution sharpening over document-topic distributions and geometric separation over topic embeddings.

\paragraph{Label-Indexed Distribution Sharpening}
For label level $l$, the document-topic distribution $\theta_d^{(l)}$ is sharpened toward the coordinate indexed by the observed label $y_d^{(l)}$, encouraging the corresponding label-indexed topic to dominate. 
We define the sharpening objective as
\begin{equation}
\mathcal{L}_{\mathrm{align}}^{(l)}
=
-\frac{1}{N}
\sum_{d=1}^{N}
\left(
1-\theta_{d,y_d^{(l)}}^{(l)}
\right)^{\gamma}
\log
\theta_{d,y_d^{(l)}}^{(l)}.
\end{equation}

Here, $\gamma$ is a focusing parameter that emphasizes low-confidence label-topic assignments.
% Here, $\gamma$ is a focusing parameter that assigns larger weights to low-confidence label-topic assignments.

\paragraph{Topic Embedding Separation}
To reduce topic collapse across labels, we further impose separation on topic embeddings.
The separation loss is defined as
\begin{equation}
\mathcal{L}_{\mathrm{sep}}^{(l)}
=
\frac{1}{K_l(K_l-1)}
\sum_{a\neq b}
\cos^2\left(T_a^{(l)},T_b^{(l)}\right).
\label{eq:sep_loss}
\end{equation}

This term penalizes high pairwise similarity among label-indexed topic embeddings without introducing an additional margin hyperparameter, thereby discouraging topic collapse.

\subsection{Dual Semantic Grounding}

Label-guided topic specialization establishes the correspondence between label indices and topic coordinates, but it does not by itself determine the semantic content of the corresponding topics. 
Since lexical and contextual representations capture complementary document information~\cite{leveraging-BERTBOW}, we ground label-specialized topics in both corpus-level lexical evidence and pretrained document semantic space.

\paragraph{BoW Reconstruction}
The bag-of-words reconstruction term grounds label-specialized topics in corpus-level lexical evidence by reconstructing the observed term-frequency vector from $\theta_d^{(l)}$ and $\beta^{(l)}$. Inspired by the use of background word frequencies in SCHOLAR~\cite{scholar}, 
we interpolate the topic-induced word distribution with a fixed global background distribution $p_{\mathrm{bg}}\in\Delta^V$, allowing common word mass to be absorbed by the background component:
\begin{equation}
\hat{x}_d^{(l)}
=
(1-\lambda_{\mathrm{bg}})
\theta_d^{(l)}\beta^{(l)}
+
\lambda_{\mathrm{bg}}p_{\mathrm{bg}}.
\label{eq:bow_recon}
\end{equation}

The reconstruction objective further weights each term by its inverse document frequency:
\begin{equation}
\mathcal{L}_{\mathrm{bow}}^{(l)}
=
-\frac{1}{N}
\sum_{d=1}^{N}
\sum_{v=1}^{V}
\omega_v x_{d,v}
\log
\hat{x}_{d,v}^{(l)},
\end{equation}
where $V$ denotes the vocabulary size, $x_{d,v}$ denotes the raw count of term $v$ in document $d$, and $\omega_v$ denotes the normalized Inverse Document Frequency (IDF) of term $v$.

\paragraph{Document Embedding Reconstruction}
In parallel, document embedding reconstruction regularizes the same topic composition in the pretrained semantic space.
We first map the topic embeddings $T^{(l)} \in \mathbb{R}^{K_l \times D_w}$ at level $l$ into the document embedding space:
\begin{equation}
U^{(l)}
=
\psi^{(l)}
\left(
T^{(l)}
\right),
\qquad
U^{(l)}\in\mathbb{R}^{K_l\times D_e}.
\end{equation}

The reconstructed document embedding is then computed as
\begin{equation}
\hat{e}_d^{(l)}
=
\theta_d^{(l)}U^{(l)}.
\label{eq:doc_recon}
\end{equation}

The document embedding reconstruction objective is defined as
\begin{equation}
\mathcal{L}_{\mathrm{doc}}^{(l)}
=
\frac{1}{N}
\sum_{d=1}^{N}
\left[
1-
\cos
\left(
\hat{e}_d^{(l)},e_d
\right)
\right].
\end{equation}

Together, the above two reconstruction objectives encourage label-specialized topics to remain grounded in corpus-specific lexical evidence and contextual document semantics.

\subsection{Hierarchical Topic Consistency}
For hierarchical labels, LGNTM encourages adjacent-level topic distributions to follow parent-child label relations using two row-normalized projection matrices:
$A_{p\rightarrow c}^{(l)} \in \mathbb{R}^{K_l \times K_{l+1}}$ from parent to child 
labels, and $A_{c\rightarrow p}^{(l)} \in \mathbb{R}^{K_{l+1} \times K_l}$ from 
child to parent labels.

For document $d$, let $\theta_d^{(l)}$ and $\theta_d^{(l+1)}$ denote its topic distributions at levels $l$ and $l+1$, respectively. The parent-to-child projection is defined as
\begin{equation}
\hat{\theta}_{d,p\rightarrow c}^{(l+1)}
=
\theta_d^{(l)}
A_{p\rightarrow c}^{(l)},
\end{equation}
while the child-to-parent projection is defined as
\begin{equation}
\hat{\theta}_{d,c\rightarrow p}^{(l)}
=
\theta_d^{(l+1)}
A_{c\rightarrow p}^{(l)}.
\end{equation}

The bidirectional hierarchical consistency objective is given by
\begin{equation}
\begin{aligned}
\mathcal{L}_{\mathrm{hier}}^{(l)}
=
\frac{1}{N}
\sum_{d=1}^{N}
\Big[
&
\mathrm{KL}
\left(
\hat{\theta}_{d,c\rightarrow p}^{(l)}
\middle\|
\theta_d^{(l)}
\right)
\\
+
&
\mathrm{KL}
\left(
\hat{\theta}_{d,p\rightarrow c}^{(l+1)}
\middle\|
\theta_d^{(l+1)}
\right)
\Big].
\end{aligned}
\end{equation}

This objective softly regularizes adjacent topic distributions to be structurally compatible with the label hierarchy.

% \subsection{Training Objective}

% The overall objective of LGNTM is defined as
% \begin{equation}
% \mathcal{L}
% =
% \sum_{l=1}^{H}
% \left(
% \mathcal{L}_{\mathrm{TS}}^{(l)}
% +
% \mathcal{L}_{\mathrm{SG}}^{(l)}
% \right)
% +
% \lambda_{\mathrm{hier}}
% \mathcal{L}_{\mathrm{HC}},
% \label{eq:final_objective}
% \end{equation}
% where
% \begin{equation}
% \begin{aligned}
% \mathcal{L}_{\mathrm{TS}}^{(l)}
% &=
% \lambda_{\mathrm{align}}
% \mathcal{L}_{\mathrm{align}}^{(l)}
% +
% \lambda_{\mathrm{sep}}
% \mathcal{L}_{\mathrm{sep}}^{(l)}, \\
% \mathcal{L}_{\mathrm{SG}}^{(l)}
% &=
% \lambda_{\mathrm{bow}}
% \mathcal{L}_{\mathrm{bow}}^{(l)}
% +
% \lambda_{\mathrm{doc}}
% \mathcal{L}_{\mathrm{doc}}^{(l)}, \\
% \mathcal{L}_{\mathrm{HC}}
% &=
% \sum_{l=1}^{H-1}
% \mathcal{L}_{\mathrm{hier}}^{(l)}.
% \end{aligned}
% \end{equation}

% Here, the $\lambda$ terms are hyperparameters. For flat labels, $H=1$ and $\mathcal{L}_{\mathrm{HC}}$ is omitted. 
% The training algorithm is summarized in Appendix~\ref{app:Training-Algorithm}.

% After training, label expansion follows Eq.~\eqref{eq:lse_topic_parameterization}, 
% and the top-$M$ expansion word set is obtained as
% \begin{equation}
% \mathcal{E}_k^{(l)}
% =
% \operatorname{TopM}_{w\in\mathcal{V}}
% \beta_k^{(l)}(w).
% \label{eq:lse_final}
% \end{equation}

\subsection{Training Objective}

The overall objective of LGNTM is defined as
\begin{equation}
\mathcal{L}
=
\sum_{l=1}^{H}
\left(
\mathcal{L}_{\mathrm{TS}}^{(l)}
+
\mathcal{L}_{\mathrm{SG}}^{(l)}
\right)
+
\lambda_{\mathrm{hier}}
\mathcal{L}_{\mathrm{HC}},
\label{eq:final_objective}
\end{equation}
where
\begin{equation}
\begin{aligned}
\mathcal{L}_{\mathrm{TS}}^{(l)}
&=
\lambda_{\mathrm{align}}
\mathcal{L}_{\mathrm{align}}^{(l)}
+
\lambda_{\mathrm{sep}}
\mathcal{L}_{\mathrm{sep}}^{(l)}, \\
\mathcal{L}_{\mathrm{SG}}^{(l)}
&=
\lambda_{\mathrm{bow}}
\mathcal{L}_{\mathrm{bow}}^{(l)}
+
\lambda_{\mathrm{doc}}
\mathcal{L}_{\mathrm{doc}}^{(l)}, \\
\mathcal{L}_{\mathrm{HC}}
&=
\sum_{l=1}^{H-1}
\mathcal{L}_{\mathrm{hier}}^{(l)}.
\end{aligned}
\end{equation}

Here, the $\lambda$ terms are hyperparameters. For flat labels, $H=1$ and
$\mathcal{L}_{\mathrm{HC}}$ is omitted.

After training, label expansion follows
Eq.~\eqref{eq:lse_topic_parameterization}, and the top-$M$ expansion word
set is obtained as
\begin{equation}
\mathcal{E}_k^{(l)}
=
\operatorname{TopM}_{w\in\mathcal{V}}
\beta_k^{(l)}(w).
\label{eq:lse_final}
\end{equation}

The complete training procedure of LGNTM is summarized in
Algorithm~\ref{alg:lgntm}. For flat label structures, the hierarchical
consistency term is omitted.

\begin{algorithm}[t]
\caption{The training procedure of LGNTM}
\label{alg:lgntm}
\begin{algorithmic}[1]
\REQUIRE
BoW matrix $X$, document semantic embeddings $E$,
hierarchical labels $Y$, and pretrained word embedding $W$.
\ENSURE
Topic-word distributions $\beta$,
hierarchical document-topic distributions $\theta$,
and label semantic expansion sets $\mathcal{E}$.

\STATE Derive levels $H$, topic numbers $K_l=|\mathcal{Y}^{(l)}|$,
label-topic correspondence $\Pi_l$, and hierarchy projections from $Y$.
\STATE Initialize the shared encoder, hierarchical decoders,
topic embeddings, and optimizer.

\WHILE{not converged}
    \FOR{each batch $(X_b,E_b,Y_b)$}
        \STATE $h \leftarrow f_{\mathrm{enc}}(E_b)$.
        \FOR{$l$ in $1:H$}
            \STATE Generate $\theta^{(l)}$ with decoder input $u^{(l)}$
            by Eqs.~\eqref{eq:decoder_input} and~\eqref{eq:theta}.
            \STATE Compute $\beta^{(l)}$ by Eq.~\eqref{eq:beta}.
            \STATE Reconstruct $\hat{X}_b^{(l)}$ and $\hat{E}_b^{(l)}$
            by Eqs.~\eqref{eq:bow_recon} and~\eqref{eq:doc_recon}.
        \ENDFOR
        \STATE Compute $\mathcal{L}$ by Eq.~\eqref{eq:final_objective}.
        \STATE Update all trainable parameters.
    \ENDFOR
\ENDWHILE

\STATE Generate label semantic expansion sets $\mathcal{E}$ from
$\beta$ by Eq.~\eqref{eq:lse_final}.
\end{algorithmic}
\end{algorithm}

\section{Experiments}
\label{sec:experiments}

We evaluate LGNTM according to the requirements of the topics-for-labels perspective. 
Since LSE relies on using label-aligned topic-word distributions as semantic representations of predefined labels, our experiments examine four questions: whether LGNTM realizes the intended one-to-one label-topic correspondence; whether the learned label-indexed topics produce corpus-grounded and interpretable label semantic expansions; whether LGNTM preserves topic quality, captures hierarchical consistency, and benefits from each component; and whether the learned expansions improve widely-adopted classification as label-side knowledge.

\subsection{Experimental Settings}
\paragraph{Datasets}

We employ three publicly available labeled datasets covering both flat and hierarchical label settings: bill summaries from the 110th--114th U.S. Congresses (2007--2017) (\textbf{Bills})~\cite{hoyle-are-NTM-broken}, medical abstracts with 5 disease-category labels (\textbf{Medical})~\cite{medical_key}, and the Web of Science dataset (\textbf{WoS})~\cite{wos_key}.
Bills and WoS use two-level label hierarchies, with 20/127 and 7/33 labels at the parent/child levels, respectively, whereas Medical uses a flat label set with 5 labels.
After removing unlabeled documents, the processed datasets contain 35,317, 11,227, and 11,913 documents for Bills, Medical, and WoS, respectively.
We preprocess the vocabulary following \citet{hoyle-are-NTM-broken}, using vocabulary sizes of 10,000 for Bills and WoS and 5,000 for Medical.

\paragraph{Baselines}
We compare LGNTM with four groups of baselines: 
1) \textbf{non-topic-modeling expansion baselines}: 
\textbf{Label Name}, which directly uses raw label names; 
\textbf{Embedding Match}, which retrieves vocabulary words nearest to each label name in a pretrained embedding space; 
and \textbf{C-TF-IDF}, which ranks words by class-based TF-IDF over label-associated documents~\cite{BERTopic};
(2) unsupervised topic models, including \textbf{LDA}~\cite{LDA}, \textbf{BERTopic}~\cite{BERTopic}, and \textbf{FASTopic}~\cite{FASTopic}; 
(3) supervised topic models, including \textbf{LLDA}~\cite{LabeledLDA}, \textbf{SCHOLAR}~\cite{scholar}, and \textbf{LANTM}~\cite{LANTM} with \textbf{ETM}~\cite{ETM} and \textbf{ECRTM}~\cite{ECRTM} backbones, denoted as \textbf{LA-ETM} and \textbf{LA-ECRTM}; 
and (4) hierarchical topic models, including \textbf{NGHTM}~\cite{NGHTM} and \textbf{TraCo}~\cite{Traco}.
For all topic-modeling methods, the number of topics at each level is set to the number of labels, i.e., $|\mathcal{Z}^{(l)}|=|\mathcal{Y}^{(l)}|=K_l$.
Other training details are provided in Appendix~\ref{app:Implementation-Details}.

\subsection{Label-Topic Alignment}
\label{sec:label-topic-alignment}
Before evaluating LSE quality, we verify whether LGNTM realizes the intended label-topic correspondence in Eq.~\eqref{eq:double-shot}.
This experiment serves as a structural diagnostic: if multiple labels are assigned to the same dominant topic, the resulting topic-word distributions cannot provide label-specific semantic expansions.

For each label $y \in \mathcal{Y}^{(l)}$, let $\mathcal{D}^{(l)}_y = \{d \mid y^{(l)}_d = y\}$ and its label-level topic distribution being
$\bar{\theta}^{(l)}_y =
|\mathcal{D}^{(l)}_y|^{-1}
\sum_{d \in \mathcal{D}^{(l)}_y}\theta^{(l)}_d$.
We define \textit{Dominant Topic Uniqueness} (DTU) at level $l$ as
\begin{equation}
\mathrm{DTU}^{(l)}
=
\frac{
\left|
\left\{
\arg\max_k \bar{\theta}^{(l)}_{y,k}
\mid
y \in \mathcal{Y}^{(l)}
\right\}
\right|
}{
K_l
}.
\label{eq:dtu}
\end{equation}

Here, $\mathrm{DTU}^{(l)}=1$ means that every label at level $l$ has a distinct dominant topic.

\begin{figure}
\centering
\begin{subfigure}{0.49\columnwidth}
\centering
\includegraphics[width=\linewidth]{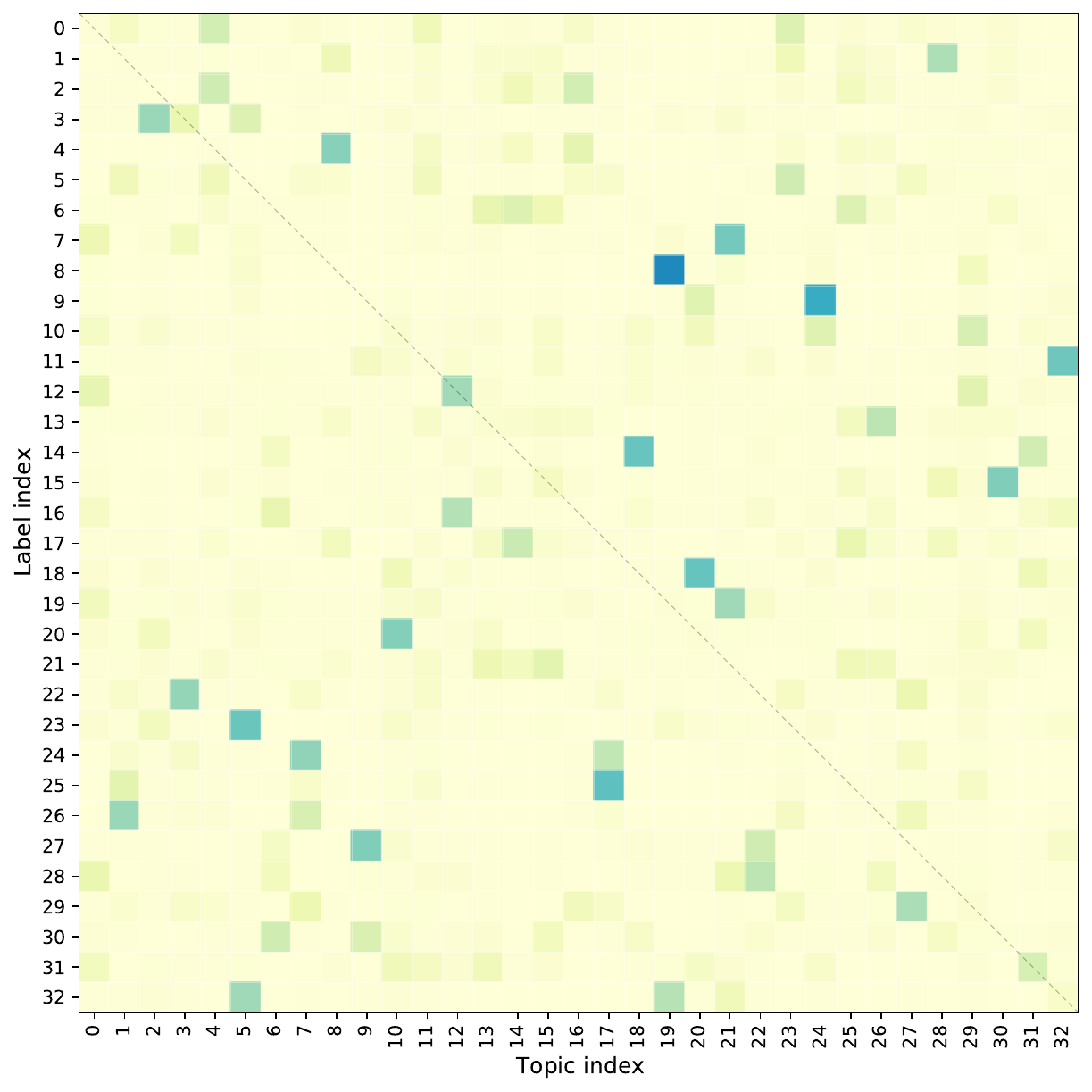}
\caption{SCHOLAR}
\end{subfigure}
\hfill
\begin{subfigure}{0.49\columnwidth}
\centering
\includegraphics[width=\linewidth]{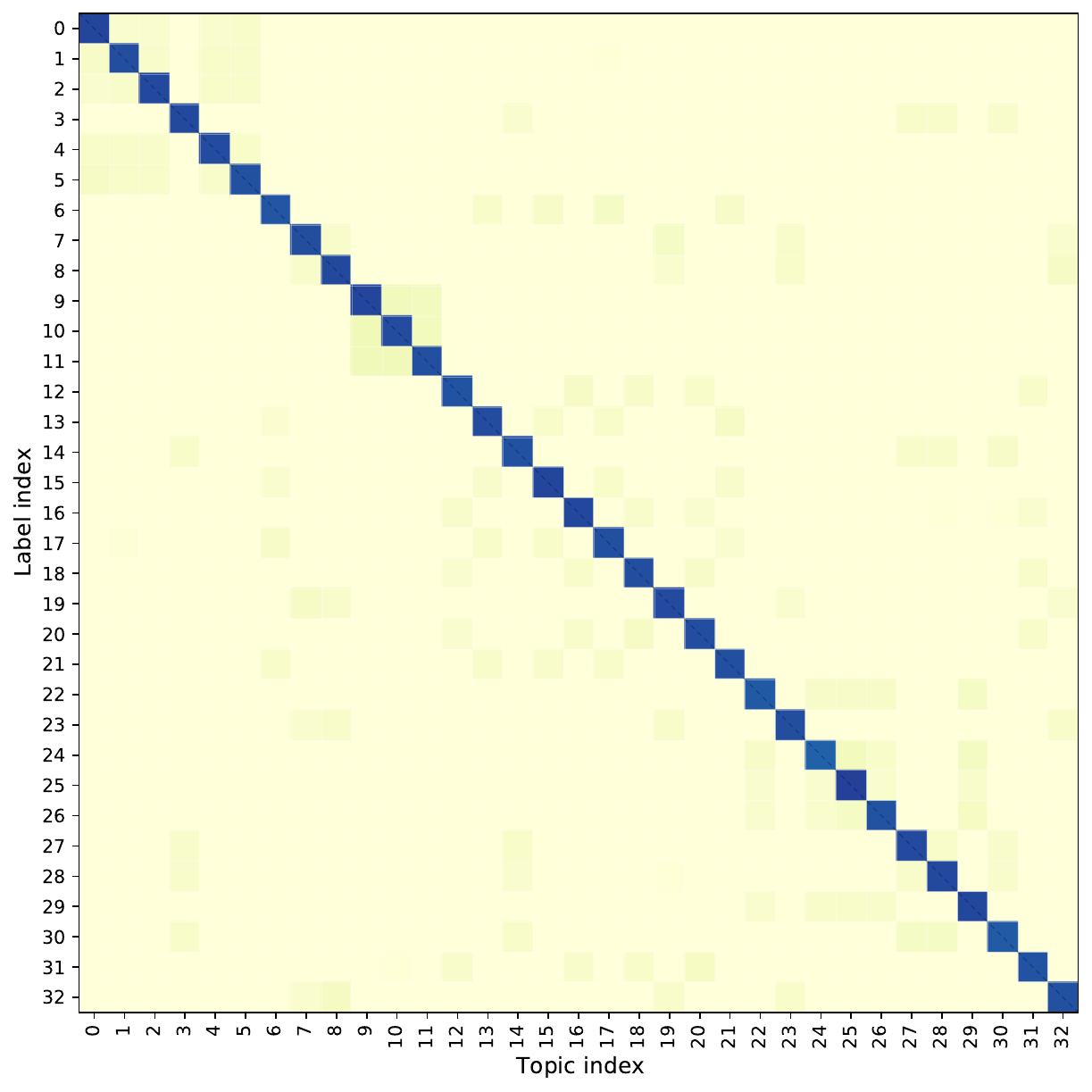}
\caption{LGNTM}
\end{subfigure}
\caption{Label-topic assignment heatmaps of SCHOLAR~\cite{scholar} and LGNTM on WoS-Child. Each cell shows the averaged topic mass $\bar{\theta}^{(l)}_{y,k}$, and the dashed diagonal marks the reference label-topic correspondence.}
\label{fig:alignment_heatmap}
\end{figure}

Full results in Appendix~\ref{app:dtu} show that LLDA and LGNTM achieve $\mathrm{DTU}^{(l)}=1.00$ across all evaluated label levels.
LLDA does so by hard-masking topic usage with observed labels, whereas LGNTM learns the correspondence through the label-guided alignment objective.
Figure~\ref{fig:alignment_heatmap} further shows that SCHOLAR maps multiple labels to shared dominant topics, while LGNTM exhibits a clear diagonal pattern.

This learned alignment provides the structural basis for using label-aligned topic-word distributions for semantic expansion.

\subsection{Label Semantic Expansion Quality}
\label{sec:semantic_expansion_quality}

\begin{table*}
\centering
\scriptsize
\setlength{\tabcolsep}{3.5pt}
\renewcommand{\arraystretch}{1.00}
\caption{Automatic evaluation of LSE, where the best and second-best results are highlighted and underlined.}
\resizebox{\textwidth}{!}{%
\begin{tabular}{lccccccccccccccc}
\toprule
\multirow{2}{*}{Model}
& \multicolumn{3}{c}{Bills-P ($K$=20)}
& \multicolumn{3}{c}{Bills-C ($K$=127)}
& \multicolumn{3}{c}{Medical ($K$=5)}
& \multicolumn{3}{c}{WoS-P ($K$=7)}
& \multicolumn{3}{c}{WoS-C ($K$=33)} \\
\cmidrule(lr){2-4}
\cmidrule(lr){5-7}
\cmidrule(lr){8-10}
\cmidrule(lr){11-13}
\cmidrule(lr){14-16}
& MAP & LNPMI & LTD
& MAP & LNPMI & LTD
& MAP & LNPMI & LTD
& MAP & LNPMI & LTD
& MAP & LNPMI & LTD \\
\midrule
Label Name
& 0.234 & -- & --
& 0.179 & -- & --
& 0.235 & -- & --
& 0.098 & -- & --
& 0.608 & -- & -- \\
Embedding Match
& 0.285 & -0.181 & 0.910
& 0.205 & -0.184 & 0.619
& \underline{0.395} & -0.214 & \underline{0.952}
& 0.216 & -0.340 & \underline{0.983}
& 0.571 & -0.164 & 0.816 \\
C-TF-IDF
& \underline{0.413} & -0.176 & \underline{0.958}
& \textbf{0.376} & 0.021 & \underline{0.684}
& 0.309 & -0.502 & \underline{0.952}
& 0.443 & -0.217 & 0.971
& \textbf{0.672} & 0.020 & \underline{0.882} \\
\midrule
LDA
& 0.281 & 0.033 & 0.286
& 0.236 & 0.051 & 0.149
& 0.325 & -0.017 & 0.373
& 0.420 & 0.025 & 0.482
& 0.501 & 0.031 & 0.298 \\
LLDA
& 0.372 & \underline{0.051} & 0.370
& \underline{0.336} & 0.055 & 0.211
& 0.392 & 0.000 & 0.528
& \textbf{0.650} & 0.030 & 0.580
& 0.590 & 0.040 & 0.360 \\
SCHOLAR
& 0.160 & -0.153 & 0.397
& 0.172 & 0.027 & 0.168
& 0.355 & \underline{0.015} & 0.699
& 0.502 & \textbf{0.085} & 0.771
& 0.589 & \textbf{0.121} & 0.639 \\
BERTopic
& 0.248 & 0.048 & 0.248
& 0.256 & \underline{0.095} & 0.246
& 0.296 & -0.016 & 0.336
& 0.360 & 0.021 & 0.375
& 0.515 & 0.071 & 0.352 \\
FASTopic
& 0.294 & \textbf{0.129} & 0.509
& 0.176 & \textbf{0.145} & 0.272
& 0.358 & \textbf{0.016} & 0.771
& 0.481 & 0.067 & 0.815
& 0.462 & \underline{0.120} & 0.567 \\
LA-ETM
& 0.199 & -0.192 & 0.340
& 0.192 & -0.209 & 0.339
& 0.312 & -0.029 & 0.563
& 0.378 & 0.028 & 0.778
& 0.344 & -0.142 & 0.257 \\
LA-ECRTM
& 0.236 & -0.119 & 0.639
& 0.136 & -0.060 & 0.265
& 0.274 & -0.144 & 0.773
& 0.370 & \underline{0.073} & 0.827
& 0.586 & 0.004 & 0.813 \\
NGHTM
& 0.130 & -0.012 & 0.102
& 0.130 & 0.003 & 0.039
& -- & -- & --
& 0.200 & -0.004 & 0.269
& 0.364 & 0.024 & 0.185 \\
TraCo
& 0.210 & 0.006 & 0.321
& 0.148 & -0.251 & 0.323
& -- & -- & --
& 0.403 & -0.083 & 0.933
& 0.597 & -0.077 & 0.875 \\
\midrule
LGNTM
& \textbf{0.449} & -0.002 & \textbf{0.988}
& 0.329 & -0.071 & \textbf{0.803}
& \textbf{0.492} & -0.019 & \textbf{1.000}
& \underline{0.611} & -0.004 & \textbf{1.000}
& \underline{0.660} & -0.042 & \textbf{0.960} \\
\bottomrule
\end{tabular}%
}
\label{tab:expansion_main}
\end{table*}

\begin{table}[t]
\centering
\scriptsize
\setlength{\tabcolsep}{2.6pt}
\renewcommand{\arraystretch}{0.90}
\caption{LLM-based evaluation of LSE quality.}
\resizebox{\columnwidth}{!}{%
\begin{tabular}{llccccc}
\toprule
Dataset & Metric & C-TF-IDF & LLDA & FASTopic & LA-ECRTM & LGNTM \\
\midrule
Bills-P
& Word Intrusion & 0.67 & 0.27 & 0.78 & \underline{0.87} & \textbf{1.00} \\
& Label Match    & \underline{0.98} & 0.95 & 0.83 & 0.75 & \textbf{1.00} \\
& Preference     & 0.02 & \underline{0.07} & \underline{0.07} & 0.00 & \textbf{0.85} \\
\midrule
Bills-C
& Word Intrusion & 0.65 & 0.33 & \underline{0.78} & 0.65 & \textbf{0.93} \\
& Label Match    & \textbf{0.98} & 0.95 & 0.95 & 0.77 & \underline{0.97} \\
& Preference     & \underline{0.13} & 0.10 & 0.05 & 0.05 & \textbf{0.67} \\
\midrule
Medical
& Word Intrusion & 0.73 & 0.20 & \underline{0.87} & 0.40 & \textbf{1.00} \\
& Label Match    & 0.87 & \underline{0.93} & 0.87 & 0.67 & \textbf{1.00} \\
& Preference     & 0.00 & \underline{0.07} & 0.00 & 0.00 & \textbf{0.93} \\
\midrule
WoS-P
& Word Intrusion & \underline{0.95} & 0.19 & 0.86 & \textbf{1.00} & \textbf{1.00} \\
& Label Match    & \underline{0.95} & 0.90 & 0.86 & 0.90 & \textbf{1.00} \\
& Preference     & 0.00 & 0.00 & \underline{0.10} & \underline{0.10} & \textbf{0.81} \\
\midrule
WoS-C
& Word Intrusion & \textbf{0.93} & 0.60 & \underline{0.81} & 0.79 & \textbf{0.93} \\
& Label Match    & \textbf{1.00} & \textbf{1.00} & \underline{0.99} & 0.92 & \textbf{1.00} \\
& Preference     & \underline{0.22} & 0.03 & 0.07 & 0.18 & \textbf{0.49} \\
\bottomrule
\end{tabular}%
}
\label{tab:expansion_llm}
\end{table}

The structural alignment alone does not guarantee useful expansion words; therefore, we next evaluate whether the induced label-conditioned word distributions benefit LSE.

To compare different methods under a unified LSE protocol, we construct an expansion word set $\mathcal{W}^{(l)}_y$ for each predefined label. 
For LGNTM, the label-topic correspondence directly gives the label-conditioned distribution as $\beta^{(l)}_y$; LLDA is handled analogously.
For topic-modeling baselines without explicit label-topic correspondence, we use the label-level topic distribution $\bar{\theta}^{(l)}_y$ to form a label-conditioned word distribution:
$\bar{\beta}^{(l)}_y = \bar{\theta}^{(l)}_y \beta^{(l)}$.
In both cases, the top-$N$ words from the corresponding label-conditioned distribution form $\mathcal{W}^{(l)}_y$.

LSE produces label-level word sets that serve as semantic descriptions of predefined labels.
Following the view that keyphrase set evaluation should be multi-aspect rather than single-dimensional~\citep{KPEval}, we evaluate LSE quality along four complementary dimensions: whether each expansion is corpus-grounded with respect to the target label, semantically coherent as a word set, distinctive from expansions of other labels, and interpretable as a label description.

For \textit{coverage}, we use the expansion word set as a BM25 query and report MAP~\cite{BM25}, which measures whether the expanded words retrieve documents belonging to the target label.
For \textit{coherence}, we adapt the topic coherence metric NPMI~\cite{lau-etal-2014-machine} to label expansion sets and report Label-level NPMI (LNPMI), together with word intrusion score (WIS)~\cite{chang2009reading}.
For \textit{distinctiveness}, we adapt topic diversity~\cite{ETM} to label expansion sets and report Label-level Topic Diversity (LTD), together with label-match accuracy, which tests whether the words identify the correct label among four candidate labels.
For \textit{interpretability}, we use a forced-choice preference test, where the judge selects the word set that best semantically expands a given label.
Following recent Large Language Model (LLM)-as-a-judge practices~\citep{GEVAL}, we use \textbf{GPT-5.1}~\cite{singh2026openaigpt5card} to perform the three judgment-based evaluations: WIS, label-match accuracy, and preference.
All LLM-based scores are averaged over three independent runs.
Tables~\ref{tab:expansion_main} and~\ref{tab:expansion_llm} report automatic and LLM-based LSE evaluations, respectively.

For coverage, LGNTM obtains the best MAP on Bills-Parent and Medical and remains competitive on the other settings. 
C-TF-IDF and LLDA achieve the highest MAP in some cases, which is expected because C-TF-IDF directly ranks class-discriminative lexical cues and LLDA hard-binds topics to observed labels. 
However, these retrieval-oriented advantages do not consistently translate into better label expansions: C-TF-IDF and LLDA obtain substantially lower preference scores, and LLDA also shows much lower LTD.
This suggests that retrieval relevance or structural label binding alone is insufficient for LSE.

For coherence and distinctiveness, several baselines such as FASTopic and SCHOLAR obtain higher LNPMI, indicating locally co-occurring word clusters. 
However, LSE requires word sets to be both internally coherent and separable across predefined labels.
LGNTM achieves the highest LTD across all five settings, showing that it produces less redundant and more label-specific expansions. 
Moreover, the LLM-based WIS and label-match results show that this distinctiveness is not arbitrary: the generated word sets remain semantically cohesive and identifiable with the intended labels. 
Finally, LGNTM substantially outperforms all baselines in the preference test, indicating that its word sets are most often judged as the best semantic descriptions of the target labels. 
% Overall, LGNTM provides the strongest LSE quality by balancing corpus grounding, semantic cohesion, cross-label distinctiveness, and interpretability.
Overall, LGNTM provides the strongest LSE quality by balancing corpus grounding, semantic cohesion, distinctiveness, and interpretability.

\subsection{Topic Model Performance}
\label{sec:topic_model_performance}

\begin{table*}[t]
\centering
\scriptsize
\setlength{\tabcolsep}{2.9pt}
\renewcommand{\arraystretch}{1.00}
% \caption{Topic model performance.}
\caption{Comparison results on topic quality and clustering performance.}
\resizebox{\textwidth}{!}{%
\begin{tabular}{lcccccccccccccccccccc}
\toprule
\multirow{2}{*}{Model}
& \multicolumn{4}{c}{Bills-Parent}
& \multicolumn{4}{c}{Bills-Child}
& \multicolumn{4}{c}{Medical}
& \multicolumn{4}{c}{WoS-Parent}
& \multicolumn{4}{c}{WoS-Child} \\
\cmidrule(lr){2-5}
\cmidrule(lr){6-9}
\cmidrule(lr){10-13}
\cmidrule(lr){14-17}
\cmidrule(lr){18-21}
& TD & $C_V$ & ARI & NMI
& TD & $C_V$ & ARI & NMI
& TD & $C_V$ & ARI & NMI
& TD & $C_V$ & ARI & NMI
& TD & $C_V$ & ARI & NMI \\
\midrule
LDA
& 0.533 & 0.522 & 0.300 & 0.417
& 0.327 & 0.593 & 0.325 & 0.559
& 0.741 & 0.536 & 0.077 & 0.116
& 0.715 & 0.503 & 0.401 & 0.477
& 0.558 & 0.609 & 0.301 & 0.548 \\
LLDA
& 0.386 & 0.464 & \underline{0.533} & \underline{0.566}
& 0.213 & 0.490 & \underline{0.464} & \underline{0.634}
& 0.504 & 0.404 & \underline{0.335} & \underline{0.352}
& 0.580 & 0.470 & \underline{0.640} & \underline{0.640}
& 0.390 & 0.550 & \underline{0.520} & \underline{0.680} \\
SCHOLAR
& 0.964 & 0.371 & 0.278 & 0.399
& 0.479 & 0.592 & 0.248 & 0.564
& \textbf{1.000} & \textbf{0.650} & 0.073 & 0.110
& 0.989 & \textbf{0.756} & 0.397 & 0.467
& 0.838 & \textbf{0.764} & 0.335 & 0.579 \\
BERTopic
& 0.750 & \textbf{0.614} & 0.272 & 0.362
& 0.612 & \textbf{0.658} & 0.321 & 0.555
& 0.662 & 0.461 & 0.203 & 0.210
& 0.710 & 0.551 & 0.374 & 0.434
& 0.721 & \underline{0.712} & 0.375 & 0.591 \\
FASTopic
& \textbf{1.000} & 0.578 & 0.275 & 0.363
& \textbf{0.903} & 0.526 & 0.257 & 0.522
& \textbf{1.000} & 0.563 & 0.133 & 0.159
& \textbf{1.000} & 0.608 & 0.414 & 0.489
& \textbf{1.000} & 0.648 & 0.326 & 0.563 \\
LA-ETM
& \underline{0.998} & 0.507 & 0.336 & 0.413
& \underline{0.870} & 0.368 & 0.285 & 0.586
& \textbf{1.000} & 0.524 & 0.084 & 0.111
& \textbf{1.000} & \underline{0.621} & 0.397 & 0.460
& 0.985 & 0.531 & 0.335 & 0.575 \\
LA-ECRTM
& 0.966 & 0.473 & 0.249 & 0.351
& 0.714 & 0.423 & 0.275 & 0.513
& \underline{0.995} & 0.440 & 0.075 & 0.095
& \underline{0.997} & 0.551 & 0.365 & 0.447
& 0.962 & 0.501 & 0.245 & 0.503 \\
NGHTM
& 0.662 & 0.500 & 0.108 & 0.187
& 0.685 & \underline{0.609} & 0.146 & 0.436
& -- & -- & -- & --
& 0.657 & 0.442 & 0.162 & 0.246
& 0.836 & 0.606 & 0.168 & 0.435 \\
TraCo
& 0.968 & 0.484 & 0.309 & 0.401
& 0.822 & 0.348 & 0.287 & 0.511
& -- & -- & -- & --
& \textbf{1.000} & 0.459 & 0.319 & 0.406
& \underline{0.996} & 0.424 & 0.320 & 0.551 \\
\midrule
LGNTM
& 0.988 & \underline{0.606} & \textbf{0.634} & \textbf{0.641}
& 0.802 & 0.494 & \textbf{0.544} & \textbf{0.679}
& \textbf{1.000} & \underline{0.572} & \textbf{0.378} & \textbf{0.390}
& \textbf{1.000} & 0.529 & \textbf{0.759} & \textbf{0.738}
& 0.955 & 0.599 & \textbf{0.583} & \textbf{0.724} \\
\bottomrule
\end{tabular}%
}
\label{tab:topic_model_performance}
\end{table*}

We further evaluate whether LGNTM remains competitive as a topic model while optimizing for LSE. 
In particular, we assess topic-word quality using topic diversity (TD) and $C_V$~\cite{Topic-Coherence-Measures}, and document-topic representation quality via ARI and NMI against ground-truth labels, following \citet{hoyle-are-NTM-broken}.

Table~\ref{tab:topic_model_performance} shows that LGNTM substantially improves document-topic representation quality, achieving the best ARI and NMI across all dataset-level blocks.
For topic-word quality, LGNTM is not uniformly best, but remains competitive in TD and $C_V$ across settings.
This indicates that optimizing for LSE does not reduce LGNTM to a purely discriminative label predictor: the model learns more label-discriminative document-topic representations while still preserving meaningful topic-word distributions.
Full hierarchical topic evaluation is provided in Appendix~\ref{app:hierarchical_evaluation}, where LGNTM shows stronger parent-child consistency on hierarchical datasets.

\subsection{Ablation Study}
% - 在 WoS/BILLS 上的表现 （选择一个数据集）： DTU + MAP + Expansion NPMI/CV +Expansion Distinctness + TD+ NPMI/CV + ARI + PCS
\label{sec:ablation}

We perform ablation experiments on WoS to validate the effectiveness of each component, removing label-indexed alignment, topic separation, lexical grounding, document grounding, and hierarchical consistency.

Table~\ref{tab:ablation} shows that label-indexed alignment and lexical grounding are the most critical components for LSE.
Removing alignment substantially reduces DTU, MAP, ARI, and PCCTG, confirming that label-indexed alignment provides the structural coordinates needed for label-centered topic representations.
Removing lexical grounding keeps DTU at 1.00 but sharply reduces MAP and LNPMI, showing that one-to-one label-topic correspondence alone is insufficient for producing useful expansion words.
Topic separation and document grounding mainly contribute to cross-label distinctiveness, as reflected by lower LTD after removing either component.
Removing hierarchical consistency primarily lowers PCCTG, confirming its role in preserving parent-child structure.

% Table~\ref{tab:ablation} shows that alignment and lexical grounding are the most critical components. Removing alignment substantially reduces DTU, MAP, ARI, and PCCTG, indicating that label-indexed alignment provides the structural basis for label-centered topic representations. Removing lexical grounding keeps DTU at 1.00 but sharply reduces MAP and LNPMI, suggesting that one-to-one correspondence alone is insufficient for producing useful expansion words. Topic separation and document grounding mainly improve cross-label distinctiveness, while hierarchical consistency primarily improves PCCTG by preserving parent-child structure.

\begin{table}
\centering
\small
\setlength{\tabcolsep}{3.6pt}
\renewcommand{\arraystretch}{0.95}
\caption{Ablation study on WoS. PCCTG denotes parent-child consistency; details are in Appendix~\ref{app:hierarchical_evaluation}.}
\begin{tabular}{lcccccc}
\toprule
Variant & DTU & MAP & LNPMI & LTD & ARI & PCCTG \\
\midrule
Full       & 1.000 & 0.641 & -0.018 & 0.979 & 0.671 & 0.422 \\
w/o Align  & 0.747 & 0.161 & -0.010 & 0.963 & 0.412 & 0.138 \\
w/o Sep.   & 1.000 & 0.635 &  0.013 & 0.951 & 0.669 & 0.417 \\
w/o Lex.   & 1.000 & 0.259 & -0.357 & 0.956 & 0.681 & 0.421 \\
w/o Doc.   & 1.000 & 0.640 &  0.012 & 0.957 & 0.666 & 0.423 \\
w/o Hier.  & 1.000 & 0.633 & -0.000 & 0.955 & 0.675 & 0.350 \\
\bottomrule
\end{tabular}
\label{tab:ablation}
\end{table}

\section{Utility Analysis}
\label{sec:utility}

We further evaluate whether the learned label expansions can serve as useful label-side semantic knowledge for LLM-based classification.
We use Qwen3-8B-AWQ~\cite{yang2025qwen3technicalreport} as the backbone LLM and evaluate on WoS-Parent and Medical, sampling 1,500 test documents from each dataset.
All variants use the same prompt template and differ only in the label-side semantic information, including label names, LLM-generated descriptions, embedding-based expansions, C-TF-IDF words, LANTM-ECRTM words, LGNTM words, and a shuffled LGNTM sanity check where LGNTM words are randomly assigned to incorrect labels.
More details are provided in Appendix~\ref{app:downstream_classification}.

\begin{table}[t]
\centering
\small
\setlength{\tabcolsep}{3pt}
\renewcommand{\arraystretch}{1.05}
\caption{LLM-based classification results.}
\begin{tabular*}{\columnwidth}{@{\extracolsep{\fill}}lcccc@{}}
\toprule
\multirow{2}{*}{Variant}
& \multicolumn{2}{c}{WoS-P}
& \multicolumn{2}{c}{Medical} \\
\cmidrule(lr){2-3}
\cmidrule(lr){4-5}
& Acc. & F1 & Acc. & F1 \\
\midrule
Label name
& 0.6367 & 0.5511 & 0.6513 & 0.5484 \\
LLM Description
& 0.6240 & 0.5403 & 0.6227 & 0.6322 \\
Embedding Match
& 0.5913 & 0.5109 & \underline{0.6560} & \underline{0.6586} \\
C-TF-IDF
& \underline{0.6847} & \underline{0.5952} & 0.6527 & 0.6568 \\
LA-ECRTM
& 0.5900 & 0.5801 & 0.6420 & 0.6481 \\
Shuffled LGNTM
& 0.4900 & 0.4021 & 0.6427 & 0.5330 \\
\textbf{LGNTM}
& \textbf{0.6920} & \textbf{0.6864}
& \textbf{0.6673} & \textbf{0.6690} \\
\bottomrule
\end{tabular*}
\label{tab:downstream_classification}
\end{table}

Table~\ref{tab:downstream_classification} shows that LGNTM achieves the best performance on both datasets, with Macro-F1 gains of 0.1353 on WoS-Parent and 0.1206 on Medical over label-name-only prompts.
The shuffled variant substantially reduces performance, confirming that the gain comes from correct label-topic-word correspondence rather than simply adding more words.
Appendix~\ref{app:case_study} provides qualitative examples showing that LGNTM produces more label-consistent expansions.

\section{Conclusion}

We introduced the \emph{topics-for-labels} perspective for enriching predefined labels with corpus-grounded descriptive topic words. To instantiate this perspective, we proposed LGNTM, a label-oriented neural topic model that learns dedicated, semantically grounded topics for predefined labels. Experiments show that the resulting label expansions are corpus-grounded, distinctive, and interpretable, while LGNTM remains competitive as a topic model. Downstream LLM-based classification further confirms the utility of these expansions as label-side semantic knowledge.

\section*{Limitations}
\label{sec:limitations}

Our work has several limitations that could be explored in future work. First, LGNTM assumes that a predefined label set or taxonomy is available, and is therefore designed for label-centered semantic expansion rather than open-ended topic discovery or automatic taxonomy induction. Second, since the generated expansion words are grounded in the training corpus, their quality and coverage may be affected by corpus distribution, label granularity, and the amount of label-associated evidence. For broad or low-resource labels, the expansions may emphasize dominant corpus-specific subtopics rather than exhaustively covering the full semantic scope of the label. Third, although LGNTM is formulated for hierarchical label structures, our experiments only cover flat or two-level label settings. Future work could further evaluate LGNTM on deeper and more complex taxonomies.

% \input{docs_new/ch6-0429}
% \input{docs/ch5-Conclusions}

% \input{docs/ch6-Limitations}

% \bibliography{custom,anthology_new}
\bibliography{custom}

\appendix

\section{Implementation Details}
\label{app:Implementation-Details}

For all datasets, we use the predefined training/test splits and average results over five runs.
The dataset statistics are shown in Table~\ref{tab:dataset_statistics}.
For flat topic models, we train one model for each label level.
For hierarchical topic models on Bills and WoS, we set the number of topic layers to $H$ and match each layer to the corresponding label level.

% \begin{table}
% \centering
% \small
% \caption{Dataset statistics. For hierarchical datasets, \#Labels denotes the number of labels at each level.}
% \begin{tabular}{lcccc}
% \toprule
% Dataset & \#Vocab & \#Train & \#Test & \#Labels  \\
% \midrule
% Bills   & 10,000 & 26,039 & 9,278 & (20, 127)  \\
% Medical & 5,000  & 8,420  & 2,807 & 5        \\
% WoS     & 10,000 & 9,531  & 2,382 & (7, 33)  \\
% \bottomrule
% \end{tabular}
% \label{tab:dataset_statistics}
% \end{table}

We use \texttt{all-MiniLM-L6-v2}~\cite{sbert} to obtain contextual document embeddings and pretrained word embeddings.
The pretrained embeddings are fixed during training.
The word embedding dimension, the document embedding dimension and the topic embedding dimension are all set to 384.
The shared encoder is implemented as a multilayer perceptron with a hidden dimension of 256.
The IDF weights $\omega_v$ in the BoW reconstruction loss are computed on the training corpus.
The global background distribution $p_{\mathrm{bg}}$ is computed from corpus-level word frequencies on the training split.

All baselines are run with public model codes.
All models are trained on a workstation equipped with an NVIDIA RTX 3090 GPU with 24 GB memory.
We use the Adam optimizer with a learning rate of $0.002$, a batch size of $200$, and $\beta=(0.99, 0.999)$.
The maximum number of training epochs is set to 400.
During training, we monitor the training objective and apply early stopping when it no longer improves.
Early stopping is activated after a warm-up period of 10 epochs, with a patience of 5 epochs.
For all baselines and LGNTM, we use the same dataset-level tuning protocol and keep the selected hyperparameters fixed across the five runs.
For each dataset, hyperparameters are selected by considering the overall performance of $C_V$, TD, and ARI.

\begin{table}
\centering
\small
\caption{Dataset statistics. For hierarchical datasets, \#Labels denotes the number of labels at each level.}
\begin{tabular}{lcccc}
\toprule
Dataset & \#Vocab & \#Train & \#Test & \#Labels  \\
\midrule
Bills   & 10,000 & 26,039 & 9,278 & (20, 127)  \\
Medical & 5,000  & 8,420  & 2,807 & 5        \\
WoS     & 10,000 & 9,531  & 2,382 & (7, 33)  \\
\bottomrule
\end{tabular}
\label{tab:dataset_statistics}
\end{table}

For Bills and WoS, the projection matrices $A_{p\rightarrow c}^{(l)}$ and $A_{c\rightarrow p}^{(l)}$ are fixed rather than learnable.
We construct them from the parent-child label co-occurrences in the training set.
For each adjacent pair of levels, we first build a binary adjacency matrix indicating whether a parent-child label pair appears in the training data.
The parent-to-child matrix $A_{p\rightarrow c}^{(l)}$ is obtained by row-normalizing this binary adjacency matrix, and the child-to-parent matrix $A_{c\rightarrow p}^{(l)}$ is obtained by row-normalizing its transpose.
Rows without observed links are assigned a uniform distribution before normalization.

For LGNTM, the focusing parameter in the label-indexed distribution sharpening loss is fixed to $\gamma=2$.
The selected LGNTM hyperparameters are reported in Table~\ref{tab:lgntm_hyperparams}.
For Medical, the hierarchical consistency term is omitted because it has a flat label structure.

\begin{table}
\centering
\small
\caption{Selected LGNTM hyperparameters on each dataset. $\lambda_{\mathrm{align}}$, $\lambda_{\mathrm{sep}}$, $\lambda_{\mathrm{bow}}$, $\lambda_{\mathrm{doc}}$, $\lambda_{\mathrm{hier}}$, and $\lambda_{\mathrm{bg}}$ denote the weights for label-indexed distribution sharpening, topic embedding separation, BoW reconstruction, document embedding reconstruction, hierarchical consistency, and background interpolation, respectively.}
\begin{tabular}{lcccccc}
\toprule
Dataset 
& $\lambda_{\mathrm{align}}$ 
& $\lambda_{\mathrm{sep}}$ 
& $\lambda_{\mathrm{bow}}$ 
& $\lambda_{\mathrm{doc}}$ 
& $\lambda_{\mathrm{hier}}$ 
& $\lambda_{\mathrm{bg}}$ \\
\midrule
Bills 
& 15.0 & 15.0 & 0.07 & 10.0 & 0.5 & 0.6 \\
WoS 
& 20.0 & 10.0 & 0.03 & 10.0 & 1.0 & 0.6 \\
Medical 
& 20.0 & 10.0 & 0.05 & 25.0 & -- & 0.5 \\
\bottomrule
\end{tabular}
\label{tab:lgntm_hyperparams}
\end{table}

\section{Dominant Topic Uniqueness}
\label{app:dtu}

Table~\ref{tab:dtu_results} reports the full Dominant Topic Uniqueness (DTU) results corresponding to Eq.~\eqref{eq:dtu}.
DTU is computed separately for each dataset-level setting.
Higher values indicate less dominant-topic sharing among labels.

% \begin{table}
% \centering
% \small
% \setlength{\tabcolsep}{3pt}
% \caption{Dominant Topic Uniqueness (DTU) results.}
% \begin{tabular}{lccccc}
% \toprule
% Model 
% & Bills-P 
% & Bills-C
% & Medical 
% & WoS-P 
% & WoS-C \\
% \midrule
% LDA 
% & 0.84 & 0.67 & 0.64 & 0.91 & 0.79 \\
% LLDA 
% & \textbf{1.00} & \textbf{1.00} & \textbf{1.00} & \textbf{1.00} & \textbf{1.00} \\
% SCHOLAR 
% & 0.77 & 0.61 & 0.76 & 0.86 & 0.82 \\
% BERTopic 
% & 0.60 & 0.60 & 0.72 & 0.83 & 0.72 \\
% FASTopic 
% & 0.64 & 0.38 & 0.68 & 0.80 & 0.71 \\
% LA-ETM 
% & 0.78 & 0.66 & 0.80 & 0.86 & 0.73 \\
% LA-ECRTM 
% & 0.46 & 0.22 & 0.72 & 0.66 & 0.47 \\
% NGHTM 
% & 0.70 & 0.54 & -- & 0.57 & 0.55 \\
% TraCo 
% & 0.73 & 0.52 & -- & 0.83 & 0.73 \\
% LGNTM 
% & \textbf{1.00} & \textbf{1.00} & \textbf{1.00} & \textbf{1.00} & \textbf{1.00} \\
% \bottomrule
% \end{tabular}
% \label{tab:dtu_results}
% \end{table}

Most baselines assign multiple labels to shared dominant topics, especially on fine-grained label spaces such as Bills-C and WoS-C.
This suggests that their topic coordinates are not consistently organized around predefined labels, limiting their direct use for label-specific semantic expansion.

LLDA and LGNTM both achieve $\mathrm{DTU}=1.00$ in all settings, but the source of this behavior differs.
LLDA obtains unique dominant topics through hard label-based topic restrictions, whereas LGNTM learns the correspondence through the label-indexed alignment objective without hard-masking topic usage.
Thus, the full DTU results support the structural role of the proposed alignment objective.

DTU should be interpreted only as a structural diagnostic.
It does not measure whether the corresponding topic-word distributions are coherent, corpus-grounded, or useful as label expansions.

\section{Label Semantic Expansion Evaluation Details}
\label{app:expansion_metrics}

\begin{table}
\centering
\small
\setlength{\tabcolsep}{3pt}
\caption{Dominant Topic Uniqueness (DTU) results.}
\begin{tabular}{lccccc}
\toprule
Model 
& Bills-P 
& Bills-C
& Medical 
& WoS-P 
& WoS-C \\
\midrule
LDA 
& 0.84 & 0.67 & 0.64 & 0.91 & 0.79 \\
LLDA 
& \textbf{1.00} & \textbf{1.00} & \textbf{1.00} & \textbf{1.00} & \textbf{1.00} \\
SCHOLAR 
& 0.77 & 0.61 & 0.76 & 0.86 & 0.82 \\
BERTopic 
& 0.60 & 0.60 & 0.72 & 0.83 & 0.72 \\
FASTopic 
& 0.64 & 0.38 & 0.68 & 0.80 & 0.71 \\
LA-ETM 
& 0.78 & 0.66 & 0.80 & 0.86 & 0.73 \\
LA-ECRTM 
& 0.46 & 0.22 & 0.72 & 0.66 & 0.47 \\
NGHTM 
& 0.70 & 0.54 & -- & 0.57 & 0.55 \\
TraCo 
& 0.73 & 0.52 & -- & 0.83 & 0.73 \\
LGNTM 
& \textbf{1.00} & \textbf{1.00} & \textbf{1.00} & \textbf{1.00} & \textbf{1.00} \\
\bottomrule
\end{tabular}
\label{tab:dtu_results}
\end{table}

We evaluate label semantic expansion using label-level word sets $\mathcal{W}^{(l)}_y$.
As described in Section~\ref{sec:semantic_expansion_quality}, LGNTM and LLDA extract words directly from the label-aligned topic-word distribution, while topic-modeling baselines without explicit label-topic correspondence use the label-conditioned distribution $\bar{\beta}^{(l)}_y=\bar{\theta}^{(l)}_y\beta^{(l)}$.
Non-topic baselines select words according to their corresponding ranking scores.

\paragraph{Automatic evaluation.}
For BM25-based MAP, the Label Name baseline uses only the raw label name as the query.
For all expansion-based methods, we concatenate the raw label name and the top-10 expansion words as the query.
Documents assigned to the target label are treated as relevant.
We compute the average precision for each label and report the macro average over labels for each dataset-level setting.

For coherence, we compute corpus-based Label-level NPMI (LNPMI) over the top-10 expansion words of each label, using word co-occurrences estimated from the training corpus.
The same top-10 setting is used for $C_v$ in Section~\ref{sec:topic_model_performance}.
For distinctiveness, we compute Label-level Topic Diversity (LTD) over the top-25 expansion words.
The same top-25 setting is used for TD in Section~\ref{sec:topic_model_performance}.
All automatic metrics are computed separately for each label level.

\paragraph{LLM-based metrics.}

All LLM-based evaluations use top-10 expansion words.
For topic-modeling methods with five runs, including LGNTM, LLDA, FASTopic, and LANTM-ECRTM, we first average the label-conditioned word distributions across runs and then extract the top-10 words.
This yields one stable word set for each label and method.

We evaluate all labels for Bills-Parent, WoS-Parent, WoS-Child, and Medical.
For Bills-Child, which contains 127 labels, we sample 20 child labels due to high evaluation costs.
The sampled subset covers all 20 Bills parent labels by selecting one child label under each parent with a fixed random seed.
All methods are evaluated on the same sampled labels.

For word intrusion, each instance consists of a target-label word set and one intruder word.
The intruder is sampled from negative-label top words and is required not to appear in the target-label word set.
The word intrusion score is the proportion of instances in which the judge correctly identifies the intruder.

For label matching, each instance consists of a word set and four candidate labels, including the gold label and three negatives.
For parent-level labels, negatives are sampled from the remaining labels.
For child-level labels, negatives are preferentially sampled from sibling labels under the same parent; if fewer than three siblings are available, the remaining negatives are sampled from other child labels.
The label-match score is the proportion of instances in which the judge selects the gold label.

For preference evaluation, each instance consists of a label and five candidate word sets generated by C-TF-IDF, LLDA, FASTopic, LANTM-ECRTM, and LGNTM.
The judge selects the word set that best semantically expands the label.
Candidate order is randomized, and each method's preference score is the proportion of instances in which its word set is selected.

We use \textbf{GPT-5.1}~\cite{singh2026openaigpt5card} as the judge for all three LLM-based tasks.
All LLM-based scores are averaged over three independent runs.
The prompt templates are shown in Figure~\ref{fig:lse_eval_prompts}.

\section{Hierarchical Evaluation}
\label{app:hierarchical_evaluation}

For hierarchical datasets, Bills and WoS, we evaluate topic hierarchy quality from two aspects: vertical consistency and non-redundancy.
\textbf{PCCTG} measures whether the child-level document-topic space preserves parent-level label grouping.
For each document $d$, we compare its cosine similarity to the centroid of its ground-truth parent group with its maximum similarity to other parent-group centroids:
\begin{equation}
\begin{aligned}
\mathrm{PCCTG}
&=
\frac{1}{|\mathcal{D}|}
\sum_{d\in\mathcal{D}}
\left[
\cos\!\left(\theta^{(c)}_d,\mu_{p(d)}\right)
\right. \\
&\qquad\left.
-
\max_{p'\neq p(d)}
\cos\!\left(\theta^{(c)}_d,\mu_{p'}\right)
\right].
\end{aligned}
\label{eq:pcctg}
\end{equation}

In the above, $\theta^{(c)}_d$ is the child-level document-topic distribution, $p(d)$ is the ground-truth parent label, and $\mu_p$ is the centroid of documents with parent label $p$.

\textbf{PCS} follows HARIN~\cite{HARIN} and measures semantic relatedness between parent and child topics using top-$K$ word-embedding similarity, with an overlap penalty for duplicated topic words.
\textbf{PCD} and \textbf{SD} follow TraCo~\cite{Traco}: PCD measures topic diversity between parent-child topic pairs, and SD measures diversity among sibling topics under the same parent.

\begin{table}
\centering
\small
\setlength{\tabcolsep}{4.5pt}
\renewcommand{\arraystretch}{0.95}
\caption{Hierarchical topic evaluation on Bills and WoS.}
\begin{tabular}{llcccc}
\toprule
Dataset & Model & PCCTG & PCS & PCD & SD \\
\midrule
Bills & NGHTM & 0.018 & 0.214 & \textbf{0.990} & \textbf{0.995} \\
      & TraCo & \underline{0.073} & 0.204 & \underline{0.987} & \underline{0.992} \\
      & LGNTM & \textbf{0.310} & \textbf{0.321} & 0.902 & 0.987 \\
\midrule
WoS   & NGHTM & 0.053 & \underline{0.257} & \underline{0.992} & 0.991 \\
      & TraCo & \underline{0.113} & 0.211 & \textbf{0.995} & \textbf{1.000} \\
      & LGNTM & \textbf{0.420} & \textbf{0.327} & 0.907 & \underline{0.993} \\
\bottomrule
\end{tabular}
\label{tab:hierarchy_quality}
\end{table}

Table~\ref{tab:hierarchy_quality} shows that LGNTM achieves the highest PCCTG and PCS on both hierarchical datasets.
This indicates stronger vertical consistency in both document-topic and topic-word spaces.
Although LGNTM does not always perform the best on PCD and SD, it remains competitive on sibling diversity, especially on WoS.
These results suggest that LGNTM preserves better hierarchical label structure than baseline methods, while maintaining reasonably distinguishable topic vocabularies.

\section{Downstream Classification Details}
\label{app:downstream_classification}

We evaluate the utility of label semantic expansions for LLM-based classification on WoS-Parent and Medical.
For each dataset, we sample 1,500 test documents according to the original label distribution, so that the sampled evaluation set preserves the class distribution of the full test split.
We report both accuracy and macro-F1.

All variants use the same classification prompt and differ only in the label-side semantic information provided for each candidate label.
For word-based variants, we use the expansion word rankings from the LSE protocol as described in Section~\ref{sec:semantic_expansion_quality}.
We lemmatize ranked words to merge inflectional variants such as \textit{vote} and \textit{voted}, remove duplicate lemmas while preserving score order, and keep the top-15 remaining words as label-side representative terms.

For the LLM Description variant, we prompt the LLM to generate one concise description for each label using only the label name and the model's parametric knowledge, as shown in Figure~\ref{fig:llm-description-generation-prompt}.
We do not provide label-associated documents when generating these descriptions.
This reflects a practical setting where all label-associated documents are often too large to fit into the context window, while using only a small subset may introduce sampling bias.
This variant therefore serves as a label-only baseline for comparing the LLM's strong general knowledge with the corpus-grounded label information mined by LGNTM.

We also include Shuffled LGNTM as a sanity check.
It uses the same LGNTM expansion word sets, but randomly reassigns them to incorrect labels through a one-to-one permutation.
Thus, every label receives one mismatched expansion set, and every expansion set is used exactly once.
This sanity check tests whether the improvement depends on the correct pairing between labels and LGNTM words; if the LLM ignores the words, shuffling them should have little effect.

We use Qwen3-8B-AWQ~\cite{yang2025qwen3technicalreport} as the backbone LLM and perform inference with the vLLM~\cite{vLLM} serving framework for all downstream classification experiments.

Figures~\ref{fig:medical-semantic-prompt} and~\ref{fig:wos-semantic-prompt} show the classification prompts for Medical and WoS-Parent, respectively.
Both prompts use the same structure: candidate labels are listed first, followed by label-side semantic information and the input document.
The placeholder \texttt{[CATEGORY\_SEMANTIC\_INFORMATION\_i]} is instantiated according to the experimental variant.
For word-based variants, it is written as \texttt{Representative terms: [WORD-SET\_i]}; for the LLM Description variant, it is written as \texttt{Description: [DESCRIPTION\_i]}.
No document-side topic hints are provided.

\begin{figure}
\centering
\begin{tcolorbox}[promptbox, title={LLM Description Generation Template}, width=\linewidth]
\small
\textbf{Input:}

{\ttfamily
You are generating a category description for a document classification label.

Task: Given one category label, write a concise and general description of what kind of documents belong to this category.

Rules: Use exactly one sentence. Follow the required sentence pattern. Keep the description between 15 and 30 words. Return JSON only.

\vspace{0.45em}
\textbf{Label:}

[LABEL\_i]
\vspace{0.45em}

\textbf{Required sentence pattern:}

This category includes documents about [field, issue, or topic], focusing on [2--4 key aspects].
\vspace{0.45em}

\textbf{Task instruction:}

Write one continuous natural-language description for the given label.
}
\vspace{0.45em}

\textbf{Output fields:}

\texttt{\{[LABEL\_i]: [DESCRIPTION\_i]\}}
\end{tcolorbox}
\caption{Prompt template for generating an LLM-based continuous description for a single label.}
\label{fig:llm-description-generation-prompt}
\end{figure}

\begin{figure*}[t]
\centering

\begin{tcolorbox}[promptbox, title={(a) Word Intrusion Template}, width=0.95\textwidth]
\small
\textbf{Input:}

{\ttfamily
You are evaluating topic-model word sets.

Task: Given a word set, identify the single word that least fits the rest of the set.

Rules: Choose exactly one intruder word. Base your judgment only on the provided words. Return JSON only.

\vspace{0.45em}
\textbf{Word set:}

[WORD\_1], [WORD\_2], ..., [WORD\_K]
\vspace{0.45em}

\textbf{Task instruction:}

Exactly one word in the set is an intruder. Select the single word that least fits the dominant meaning of the other words.
}
\vspace{0.45em}

\textbf{Output fields:}

\texttt{\{intruder\_word\}}
\end{tcolorbox}

\vspace{0.8em}

\begin{tcolorbox}[promptbox, title={(b) Label Matching Template}, width=0.95\textwidth]
\small
\textbf{Input:}

{\ttfamily
You are evaluating whether a word set matches one of several candidate labels.

Task: Given a word set and several candidate labels, choose the single label that is best supported by the words.

Rules: Choose exactly one label. Base your judgment only on the provided words and candidate labels. Return JSON only.

\vspace{0.45em}
\textbf{Word set:}

[WORD\_1], [WORD\_2], ..., [WORD\_K]
\vspace{0.45em}

\textbf{Candidate labels:}

A. [LABEL\_A]

B. [LABEL\_B]

C. [LABEL\_C]

D. [LABEL\_D]
\vspace{0.45em}

\textbf{Task instruction:}

Choose the single candidate label that is best matched by the word set.
}
\vspace{0.45em}

\textbf{Output fields:}

\texttt{\{chosen\_label\}}
\end{tcolorbox}

\vspace{0.8em}

\begin{tcolorbox}[promptbox, title={(c) Label-Word-Set Preference Template}, width=0.95\textwidth]
\small
\textbf{Input:}

{\ttfamily
You are evaluating label semantic expansion word sets.

Task: Given a label and five candidate word sets, choose the single word set that best matches and semantically expands the label.

Rules: Choose exactly one word set. Base your judgment only on the label and the provided candidate word sets. Prefer the set that is more relevant, coherent, distinctive, and interpretable. Return JSON only.

\vspace{0.45em}
\textbf{Label:}

[LABEL]
\vspace{0.45em}

\textbf{Candidate word sets:}

Set A: [WORD\_A1], [WORD\_A2], ..., [WORD\_AK]

Set B: [WORD\_B1], [WORD\_B2], ..., [WORD\_BK]

Set C: [WORD\_C1], [WORD\_C2], ..., [WORD\_CK]

Set D: [WORD\_D1], [WORD\_D2], ..., [WORD\_DK]

Set E: [WORD\_E1], [WORD\_E2], ..., [WORD\_EK]
\vspace{0.45em}

\textbf{Task instruction:}

Choose the single candidate word set that best matches and semantically expands the label.
}
\vspace{0.45em}

\textbf{Output fields:}

\texttt{\{winner\_set\}}
\end{tcolorbox}

\caption{Prompt templates for LLM-based evaluation: (a) word intrusion, (b) label matching, and (c) label-word-set preference.}
\label{fig:lse_eval_prompts}
\end{figure*}

\begin{figure}[t]
\centering
\begin{tcolorbox}[promptbox, title={MEDICAL Classification Template}, width=\linewidth]
\small
\textbf{Input:}

{\ttfamily
You are a Medical Specialist \& Clinical Researcher. You are working with the Medical Abstract Dataset containing patient-related clinical summaries.

Your task is to determine the correct medical category for this document based on the available candidate categories.

There are \textbf{5} candidate categories in this task, and you must select exactly one.

\vspace{0.45em}
\textbf{\#\#\# Candidate Categories:}

- [LABEL\_1]

- [LABEL\_2]

...

- [LABEL\_5]

\vspace{0.45em}
\textbf{\#\#\# Category Semantic Information:}

The following information provides auxiliary semantic guidance for interpreting each category.

\vspace{0.45em}
- [LABEL\_i]

\hspace{1em}[CATEGORY\_SEMANTIC\_INFORMATION\_i]

\vspace{0.45em}
\textbf{\#\#\# Medical Abstract Text:}

[TEXT]

\vspace{0.45em}
\textbf{\#\#\# Instruction:}

1. Read the document carefully and compare it against the candidate categories.

2. Use the Category Semantic Information as clues to understand the semantic field and typical content associated with each category.

3. Base your final decision primarily on the document content as a whole.

4. Return only a valid JSON object with fields ``analysis'' and ``prediction''.

\vspace{0.3em}
\textbf{\#\#\# Output Format:}

\{\\
\hspace*{1em}``analysis'': ``your reasoning text'',\\
\hspace*{1em}``prediction'': ``one label from Candidate Categories''\\
\}

\vspace{0.45em}

\vspace{0.45em}
\textbf{\#\#\# Response:}
}
\end{tcolorbox}
\caption{Prompt template for the Medical classification task.}
\label{fig:medical-semantic-prompt}
\end{figure}

\begin{figure}[t]
\centering
\begin{tcolorbox}[promptbox, title={WoS Classification Template}, width=\linewidth]
\small
\textbf{Input:}

{\ttfamily
You are a Multidisciplinary Academic Research Analyst. You are working with the Web of Science Academic Abstract Dataset (WoS) containing multidisciplinary paper abstracts.

Your task is to determine the correct academic category for this abstract based on the available candidate categories.

There are \textbf{7} candidate categories in this task, and you must select exactly one.

\vspace{0.45em}
\textbf{\#\#\# Candidate Categories:}

- [LABEL\_1]

- [LABEL\_2]

...

- [LABEL\_7]

\vspace{0.45em}
\textbf{\#\#\# Category Semantic Information:}

The following information provides auxiliary semantic guidance for interpreting each category.

\vspace{0.45em}
- [LABEL\_i]

\hspace{1em}[CATEGORY\_SEMANTIC\_INFORMATION\_i]

\vspace{0.45em}
\textbf{\#\#\# Academic Abstract Text:}

[TEXT]

\vspace{0.45em}
\textbf{\#\#\# Instruction:}

1. Read the document carefully and compare it against the candidate categories.

2. Use the Category Semantic Information as clues to understand the semantic field and typical content associated with each category.

3. Base your final decision primarily on the document content as a whole.

4. Return only a valid JSON object with fields ``analysis'' and ``prediction''.

\vspace{0.3em}
\textbf{\#\#\# Output Format:}

\{\\
\hspace*{1em}``analysis'': ``your reasoning text'',\\
\hspace*{1em}``prediction'': ``one label from Candidate Categories''\\
\}

\vspace{0.45em}
\textbf{\#\#\# Response:}
}
\end{tcolorbox}
\caption{Prompt template for the WoS-Parent classification task.}
\label{fig:wos-semantic-prompt}
\end{figure}

\section{Qualitative Case Study}
\label{app:case_study}

\begin{table*}[t]
\centering
\scriptsize
\setlength{\tabcolsep}{0pt}
\renewcommand{\arraystretch}{1.04}
\caption{Top-15 label expansions on WoS-Parent and Medical. Each label is compared across C-TF-IDF, Labeled LDA, FASTopic, LANTM-ECRTM, and LGNTM.}
\resizebox{\textwidth}{!}{%
\begin{tabular}{@{}
>{\raggedright\arraybackslash}p{1.45cm}
@{\hspace{4pt}}
l
@{\hspace{8pt}}
l
@{}}
\toprule
Label & Model & Top-15 Words \\
\midrule

\multicolumn{3}{@{}l}{\textbf{WoS-Parent}} \\
\midrule
\multirow{5}{1.45cm}{\textit{CS (Computer Science)}}
& \mbox{C-TF-IDF} & cryptography, encryption, cryptographic, cipher, linux, ecc, secret, attacker, malware, sdn, iot, malicious, password, hash, authentication \\
& \mbox{LLDA} & use, system, network, propose, base, method, model, datum, security, paper, result, algorithm, image, application, feature \\
& \mbox{FASTopic} & computer, detection, security, algorithm, attack, network, image, software, manufacturing, sensor, mobile, smart, platform, real, ambient \\
& \mbox{LA-ECRTM} & discrete, digital, controller, simulink, avalanche, prototype, circuit, topology, computation, microgrid, realize, filtering, electromagnetic, mpc, chaotic \\
& \mbox{LGNTM} & security, detection, cryptography, recognition, cryptographic, algorithm, computing, privacy, protocol, encryption, implementation, networking, secure, surveillance, classification \\
\cmidrule(lr){1-3}

\multirow{5}{1.45cm}{\textit{Medical}}
& \mbox{C-TF-IDF} & allergy, spondylitis, asthma, allergic, ankylosing, allergen, tnf, amyloid, spondyloarthritis, axspa, ige, serum, addiction, opioid, peanut \\
& \mbox{LLDA} & patient, study, use, disease, result, treatment, allergy, group, effect, level, high, control, associate, increase, clinical \\
& \mbox{FASTopic} & patient, child, participant, clinical, cognitive, social, disorder, allergy, age, symptom, brain, anxiety, adult, emotion, disease \\
& \mbox{LA-ECRTM} & beta, alpha, tnf, inflammatory, necrosis, igg, anti, colitis, steroid, skin, systemic, bowel, sera, lupus, rash \\
& \mbox{LGNTM} & alzheimer, adhd, asthma, allergy, addiction, arthritis, disorders, allergic, allergies, disease, anxiety, dementia, chronic, inflammation, illness \\
\cmidrule(lr){1-3}

\multirow{5}{1.45cm}{\textit{ECE (Electrical Engineering)}}
& \mbox{C-TF-IDF} & converter, inverter, anode, electricity, inductor, controller, mfc, teg, resonant, inductance, pll, buck, pwm, capacitor, amplifier \\
& \mbox{LLDA} & system, control, use, power, circuit, energy, propose, model, paper, result, current, time, electrical, base, method \\
& \mbox{FASTopic} & voltage, circuit, electrical, converter, fluid, controller, mechanic, electric, finite, rotor, transient, magnet, loop, torque, generator \\
& \mbox{LA-ECRTM} & discrete, electromagnetic, verify, digital, circuit, simulink, controller, output, avalanche, orthogonal, generator, antenna, converter, mode, ghz \\
& \mbox{LGNTM} & electricity, circuit, electrical, converter, voltage, inverter, electric, electrically, current, power, electronic, energy, capacitor, photovoltaic, piezoelectric \\
\cmidrule(lr){1-3}

\multirow{5}{1.45cm}{\textit{MAE (Mechanical Engineering)}}
& \mbox{C-TF-IDF} & cad, torque, magnet, rotor, hydraulic, reynolds, stator, cam, winding, vortex, xylem, scour, fea, floodplain, neutronic \\
& \mbox{LLDA} & design, use, model, flow, machine, result, system, method, study, base, paper, process, present, high, analysis \\
& \mbox{FASTopic} & voltage, circuit, electrical, converter, fluid, controller, mechanic, finite, electric, rotor, transient, magnet, loop, numerical, torque \\
& \mbox{LA-ECRTM} & electromagnetic, discrete, output, generator, verify, circuit, antenna, power, simulink, orthogonal, avalanche, digital, finite, controller, ghz \\
& \mbox{LGNTM} & mechanics, mechanical, hydraulic, cylinder, aerodynamic, engineering, manufacturing, simulation, inlet, machining, stokes, motor, flow, pneumatic, engineer \\

\midrule
\multicolumn{3}{@{}l}{\textbf{Medical}} \\
\midrule

\multirow{5}{1.45cm}{\raggedright\textit{digestive system diseases}}
& \mbox{C-TF-IDF} & coeliac, gallbladder, pylori, lithotripsy, gallstone, capd, irritable, gastroesophageal, rehydration, cryptosporidium, ursodeoxycholic, eis, volvulus, peptic, hernia \\
& \mbox{LLDA} & patient, disease, study, group, liver, treatment, increase, result, case, use, control, cell, gastric, year, ulcer \\
& \mbox{FASTopic} & metastasis, duct, carcinoma, cyst, radiation, tumour, bladder, endoscopic, neoplasm, benign, metastatic, lymph, adenocarcinoma, histologic, malignant \\
& \mbox{LA-ECRTM} & hematopoietic, syngeneic, egf, cytotoxic, retinoic, nude, immunoassay, murine, hbc, mononuclear, immunologic, promyelocytic, effector, trans, anti \\
& \mbox{LGNTM} & bile, bowel, stomach, gallbladder, liver, gastric, gastrointestinal, hepatic, cirrhosis, colitis, ulcer, intestinal, gastroenteritis, abdominal, pancreatitis \\
\cmidrule(lr){1-3}

\multirow{5}{1.45cm}{\textit{neoplasms}}
& \mbox{C-TF-IDF} & melanoma, osteosarcoma, igf, lak, mammography, etoposide, endometrial, oncogene, mastectomy, mammographic, melanomas, carboplatin, glioma, papanicolaou, sclc \\
& \mbox{LLDA} & patient, cell, tumor, cancer, carcinoma, case, study, year, treatment, disease, use, group, high, result, human \\
& \mbox{FASTopic} & metastasis, carcinoma, duct, tumour, cyst, bladder, lymph, adenocarcinoma, radiation, neoplasm, endoscopic, metastatic, squamous, benign, melanoma \\
& \mbox{LA-ECRTM} & hematopoietic, nude, murine, lymphocyte, retinoic, egf, syngeneic, effector, ctl, immunoassay, promyelocytic, oncogene, haplotype, cytotoxic, colony \\
& \mbox{LGNTM} & tumor, cancer, tumour, carcinoma, malignancy, carcinomas, malignant, neoplasia, oncology, leukemia, metastasis, adenocarcinoma, tumorigenesis, cytologic, carcinogenesis \\
\cmidrule(lr){1-3}

\multirow{5}{1.45cm}{\raggedright\textit{nervous system diseases}}
& \mbox{C-TF-IDF} & cbz, tourette, vpa, spasticity, epileptic, achr, jakob, creutzfeldt, tic, neuroleptic, baclofen, fns, gilles, fragile, parkinson \\
& \mbox{LLDA} & patient, study, group, use, disease, control, brain, result, year, case, pain, treatment, clinical, cerebral, effect \\
& \mbox{FASTopic} & nerve, spinal, motor, seizure, cord, neurologic, sleep, headache, temporal, visual, deficit, movement, epidural, pain, nervous \\
& \mbox{LA-ECRTM} & resistance, norepinephrine, shr, phenylephrine, lactate, vasodilator, halothane, reperfusion, rat, wky, adrenoceptor, angiotensin, constant, thromboxane, nitroprusside \\
& \mbox{LGNTM} & neurological, cerebral, brain, nerve, dementia, eeg, neuronal, neuralgia, stroke, neuropsychological, neurologic, nervous, headache, impairment, neuroleptic \\
\cmidrule(lr){1-3}

\multirow{5}{1.45cm}{\raggedright\textit{cardiovascular diseases}}
& \mbox{C-TF-IDF} & doxazosin, angioplasty, shr, ptca, ami, captopril, tachycardia, transluminal, chf, sestamibi, atenolol, chd, reinjection, antihypertensive, cardiomyopathy \\
& \mbox{LLDA} & patient, group, coronary, pressure, study, artery, blood, heart, increase, ventricular, disease, myocardial, use, effect, year \\
& \mbox{FASTopic} & artery, ventricular, coronary, complication, myocardial, cardiac, surgery, procedure, right, infarction, valve, left, postoperative, aortic, graft \\
& \mbox{LA-ECRTM} & min, minute, contractility, oxygen, maximal, nitroprusside, cardioplegia, vasodilator, peak, pressure, hypertonic, conscious, slope, cardioplegic, volume \\
& \mbox{LGNTM} & cardiovascular, coronary, hypertension, cardiac, artery, myocardial, arterial, heart, systolic, circulation, endocardial, angiotensin, diastolic, echocardiography, infarction \\

\bottomrule
\end{tabular}%
}
\label{tab:case_unified}
\end{table*}

Table~\ref{tab:case_unified} provides qualitative examples that help explain the LSE evaluation results in Tables~\ref{tab:expansion_main} and~\ref{tab:expansion_llm}, as well as the downstream classification results in Table~\ref{tab:downstream_classification}.

\paragraph{Discriminative lexical cues in C-TF-IDF.}
C-TF-IDF serves as a discriminative lexical baseline: it merges all documents associated with each label into a class-level document, and ranks words according to their class-level term frequency normalized by their prevalence across label-level documents.
This design naturally favors words that are useful for retrieving target-label documents, which explains its competitive MAP scores.
However, LSE requires more than retrieval cues.
For coarse parent labels, highly discriminative words may come from only one child category, or from rare technical contexts, and therefore fail to describe the semantic scope of the parent label.

For example, in WoS-Parent, the Medical parent label contains five child labels: Addiction, Allergies, Alzheimer's Disease, Ankylosing Spondylitis, and Anxiety.
C-TF-IDF extracts terms such as \textit{axspa}, referring to axial spondyloarthritis, and \textit{ige}, referring to immunoglobulin E in allergic responses.
These terms are medically meaningful and can help retrieve documents from specific child categories, but they overrepresent narrow child-level phenomena and provide limited coverage of the broader parent-level scope of Medical.
Similarly, for MAE, terms such as \textit{xylem} and \textit{neutronic} are domain-specific and do not by themselves characterize the central semantic scope of mechanical engineering.

In contrast, LGNTM produces broader parent-level descriptions, covering multiple child-level subcategories rather than emphasizing only local discriminative terms.
This helps explain why C-TF-IDF is competitive in MAP but weaker than LGNTM in the downstream LLM-based classification task, where its WoS-Parent Macro-F1 is 0.5952 compared with 0.6864 for LGNTM.

\paragraph{Generic but coherent words in LLDA.}
LLDA illustrates a different limitation.
Although LLDA uses hard label-based masking so that each label is associated with a unique dominant topic, its word distributions often contain generic corpus-level words.

For example, in the Medical dataset, LLDA repeatedly produces high-frequency medical-corpus words such as \textit{patient}, \textit{study}, \textit{group}, \textit{use}, \textit{disease}, \textit{result}, and \textit{treatment} across different disease labels.
These words frequently occur in recurring clinical-abstract contexts, which helps LLDA remain competitive in MAP and obtain consistently higher scores than LGNTM on the co-occurrence-based LNPMI metric.

However, such locally coherent words are often generic across labels and provide limited value as label semantic expansions.
This limitation is reflected by LLDA's much lower LTD and LLM-based word intrusion scores.
Thus, LLDA shows that high LNPMI and structural label-topic binding do not by themselves guarantee useful LSE; useful label expansions also require label-specific and distinctive topic words.

\paragraph{Label drift in FASTopic and LA-ECRTM.}
The results of FASTopic and LA-ECRTM further show that local topic coherence or soft label awareness does not necessarily produce useful label expansions.
FASTopic often forms coherent word clusters, which is consistent with its strong LNPMI in several settings, but these clusters may correspond to neighboring labels rather than the target label.
LA-ECRTM uses soft label-topic indicators to encourage label-topic association, but dedicated label-specific topic-word distributions are still not guaranteed.

A common example appears in WoS-Parent MAE: FASTopic assigns ECE-related words such as \textit{voltage}, \textit{circuit}, \textit{electrical}, and \textit{converter}, while LA-ECRTM also drifts toward electrical/control terminology such as \textit{electromagnetic}, \textit{circuit}, \textit{antenna}, \textit{power}, \textit{simulink}, and \textit{controller}.
These word sets are locally meaningful, but they do not faithfully describe mechanical engineering.

In contrast, LGNTM produces MAE words such as \textit{mechanics}, \textit{mechanical}, \textit{hydraulic}, \textit{engineering}, \textit{manufacturing}, \textit{simulation}, \textit{machining}, and \textit{flow}.
This explains why FASTopic and LA-ECRTM can obtain reasonable coherence or word-intrusion scores, while their preference scores remain much lower than LGNTM.

\paragraph{Overall interpretation.}
In summary, the aforementioned examples show that LGNTM better balances corpus grounding, label-level semantic coverage, and cross-label distinctiveness.
They also illustrate why LSE quality should be evaluated from multiple complementary dimensions rather than by a single metric.

C-TF-IDF can achieve competitive MAP by extracting strong discriminative retrieval cues, but these cues may be overly narrow or noisy.
LLDA can obtain structurally aligned topics, but its expansions may contain generic high-frequency words.
FASTopic can produce locally coherent word clusters, and LA-ECRTM can encourage label-topic association, but their word sets may still drift toward neighboring labels.
In contrast, LGNTM produces expansions that are more consistently corpus-grounded, label-specific, distinctive, and interpretable.

Since all downstream variants use the same prompt template and differ only in label-side semantic information, these qualitative patterns help explain why LGNTM improves downstream LLM-based classification while Shuffled LGNTM degrades performance.
The gain therefore comes from the correct pairing between labels and corpus-grounded expansion words, rather than from simply adding more words to the prompt.

\end{document}